\documentclass[sigconf]{acmart}
\usepackage{caption}
\usepackage{subcaption}
\usepackage{algorithmic}
\usepackage[ruled,vlined]{algorithm2e}
\usepackage{adjustbox}
\AtBeginDocument{%
  \providecommand\BibTeX{{%
    \normalfont B\kern-0.5em{\scshape i\kern-0.25em b}\kern-0.8em\TeX}}}

\acmDOI{10.1145/nnnnnnn.nnnnnnn}

\acmISBN{978-x-xxxx-xxxx-x/YY/MM}

\acmConference[GECCO '22]{the Genetic and Evolutionary Computation Conference 2022}{July xx--xx, 2022}{Boston, USA}
\acmYear{2022}
\copyrightyear{2022}

\acmPrice{15.00}

\begin{document}

\title{Novel ensemble collaboration method for dynamic scheduling problems}

\author{Marko Đurasević}
\affiliation{%
  \institution{University of Zagreb, Faculty of Electrical Engineering and Computing}
\city{Zagreb} 
\country{Croatia} 
}
\email{marko.durasevic@fer.hr}

\author{Lucija Planinić}
\affiliation{%
  \institution{University of Zagreb, Faculty of Electrical Engineering and Computing}
\city{Zagreb} 
\country{Croatia} 
}
\email{lucija.planinic@fer.hr}

\author{Francisco Javier Gil Gala}
\affiliation{%
  \institution{University of Oviedo. Department of Computing}
  \city{Gijón}
  \country{Spain}}
\email{giljavier@uniovi.es}

\author{Domagoj Jakobović}
\affiliation{%
  \institution{University of Zagreb, Faculty of Electrical Engineering and Computing}
\city{Zagreb} 
\country{Croatia} 
 }
\email{domagoj.jakobovic@fer.hr}

%\renewcommand{\shortauthors}{Đurasević, et al.}

\begin{abstract}
Dynamic scheduling problems are important optimisation problems with many real-world applications. Since in dynamic scheduling not all information is available at the start, such problems are usually solved by dispatching rules (DRs), which create the schedule as the system executes. Recently, DRs have been successfully developed using genetic programming. However, a single DR may not efficiently solve different problem instances. Therefore, much research has focused on using DRs collaboratively by forming ensembles. In this paper, a novel ensemble collaboration method for dynamic scheduling is proposed. In this method, DRs are applied independently at each decision point to create a simulation of the schedule for all currently released jobs. Based on these simulations, it is determined which DR makes the best decision and that decision is applied. The results show that the ensembles easily outperform individual DRs for different ensemble sizes. Moreover, the results suggest that it is relatively easy to create good ensembles from a set of independently evolved DRs.
\end{abstract}

\begin{CCSXML}
<ccs2012>
   <concept>
       <concept_id>10010147.10010178.10010199</concept_id>
       <concept_desc>Computing methodologies~Planning and scheduling</concept_desc>
       <concept_significance>500</concept_significance>
       </concept>
 </ccs2012>
\end{CCSXML}

\ccsdesc[500]{Computing methodologies~Planning and scheduling}

\keywords{Genetic Programming, Scheduling, Unrelated Machines, Dispatching Rules, Ensembles}


\maketitle

\section{Introduction}

Scheduling is an essential part of many real-world processes, which are directly affected by the quality of the schedules created. As a result, much of the research has focused on studying various scheduling problems and methods to achieve the best possible results. Most scheduling problems are NP-hard, which makes it difficult to obtain optimal solutions in a reasonable time \cite{Pinedo2012}. As a result, scheduling problems are usually solved using various heuristic methods. However, many scheduling problems are also dynamic in nature, which means that not all problem properties are known at the beginning, but become available only during the execution of the system. This complicates the application of standard search based methods, such as genetic algorithms. Dispatching rules (DRs) have therefore become the method of choice for dynamic scheduling problems.

DRs are simple heuristics that iteratively construct the schedule \citep{DURASEVIC2018c}. At each decision point, they only determine which job should be executed next. Thus, as new jobs are released into the system or other changes occur, these rules can easily adapt to such situations. Since manual development of good DRs is difficult, research has been done on automatic development of DRs in recent years. In these studies, genetic programming (GP) \cite{Poli2008} has emerged as the method of choice for automatic development of new DRs, as it was demonstrated that the rules generated by GP outperform the best manually developed DRs in most cases.

However, there is a limit to how good results only a single DR can achieve. This is a direct consequence of the No Free Lunch theorem \cite{Wolpert1997}, which states that no one method can achieve the best results on all problems. Therefore, it is also impossible for GP to develop a single DR that will produce good results on all problem instances it may encounter. Many studies have focused on improving the performance of individual DRs \cite{Nguyen2019, GILGALA2021a}, but such a rule will still perform poorly on some instances. Therefore, several studies have focused on investigating methods for multiple DRs to collaborate in solving a problem \cite{Park2015, Durasevic2017, GilGala2019}. The motivation for such a research direction is that if a single DR cannot always make the best decisions, perhaps a collection of DRs making decisions together can produce better results.
 
Most studies that addressed the idea of using multiple DRs together used them in the form of ensembles. Although various ensemble learning methods have been proposed and used in the literature \cite{Durasevic2017, GilGala2020a}, all studies showed that ensembles of DRs performed significantly better than individual DRs. This in turn motivated several follow-up studies that looked more deeply into the problem of designing ensembles.

In this study, we explore a slightly different way in which DRs forming an ensemble can work together, which differs in some respects from the standard ensembles in the literature. More specifically, the DRs contained in the ensemble work together to solve a problem instance, but unlike standard ensembles where the independent decisions of all rules are aggregated, in these ensembles the decisions of all the rules are simulated individually. This means that each time a decision needs to be made, all the rules in the ensemble are evaluated and their decisions are assessed. The best decision is then applied to the schedule. In this way, the ensemble tries to select the best available decision and apply it to the schedule. The specific objectives of this study can be summarised as follows:
\begin{enumerate}
	\item Design a novel approach in which the DRs in an ensemble work together.
	\item Construct ensembles of DRs and apply the proposed collaboration approach to study their performance.
	\item Analyse various aspects of the proposed ensemble collaboration method.
\end{enumerate}

The remainder of the paper is organised as follows. Section \ref{sec:relatedwork} provides a literature review. Section \ref{sec:background} describes the description on the unrelated machines environment, and the application of GP to develop DRs for this environment. The proposed ensemble method is described in Section \ref{sec:collab}. Section \ref{sec:exp} describes the experimental setup and outlines the results obtained. The conclusion and future work directions are outlined in Section \ref{sec:conc}.

\section{Related work}
\label{sec:relatedwork}
From the first works that focused on the automatic generation of DRs for different environments such as single machines \cite{DIMOPOULOS2001} and job shop \citep{Miyashita2000, Jakobovic2006}, a large number of studies in this area have been published to date \cite{Branke2016, Nguyen2017}. Several areas of automated dispatching rule design have been extensively researched, such as multi-objective optimisation \cite{Nguyen2013, Nguyen2015, Durasevic2017, Zhang2019}, application of surrogate models \cite{Nguyen2017a, Zhang2021c}, and feature selection \cite{Zhang2019d, Zhang2021d}. In recent years, there have also been some new research directions, which include multitask GP \cite{Zhang2021, Zhang2021a}, improving selection mechanisms in algorithms and operators \cite{Masood2020, Zhang2020c, Planinic2021}, considering additional scheduling properties \cite{Park2018a, GilGala2019, JAKLINOVIC2021}, and generating initial populations of a genetic algorithm with automatically generated DRs \cite{VLASIC2019}.

In addition to the above research directions, the creation of ensembles of DRs represents another goal that has been intensively studied in the literature. One of the first studies on the creation of ensembles of DRs for the static job shop environment was conducted in \cite{Park2015}. The authors use the cooperative coevolution algorithm. The algorithm contains the same number of subpopulations as there are elements in the ensemble, and each subpopulation is used to evolve a single DR for the ensemble. The DRs of the individual populations interact only when evaluated as an ensemble. In \cite{Park2016}, the authors apply a multilevel genetic programming method to evolve ensembles. This method did not achieve better results than the method of \cite{Park2015}, but it evolved ensembles in less time. In \cite{Hart2016}, the authors propose NELLI-GP, an ensemble learning method for generating heuristic ensembles. The results show that the proposed method performs better than previous approaches. In \cite{Park2017}, the authors investigate different ensemble combination schemes, including linear combination, weighted linear combination, majority voting, and weighted majority voting. The results show that the linear combination scheme generally achieves the best performance. %In addition, the weighted variants of the combination methods were also shown to perform poorly and were often biased in favour of certain members of the ensemble.
 
In \cite{Durasevic2017a}, the authors compared 4 methods for constructing ensembles of DRs for the unrelated machines environment, including BagGP, BoostGP, cooperative coevolution, and the simple ensemble combination (SEC) method. The results showed that most of the obtained ensembles performed significantly better than individual DRs. Additional analysis revealed that the best results were obtained by ensembles with smaller size. For all tested methods, SEC mostly created the best ensembles. Since SEC is a simpler method compared to the other tested approaches, this motivated the authors to perform a more thorough analysis of the method, especially considering different ways to construct the ensembles \cite{Durasevic2019}. This study also showed that, even when using different ensemble construction methods, it was possible to obtain good ensembles in a short time. The ensemble approaches from \cite{Durasevic2017} were also applied to the resource constrained project scheduling problem (RCPSP) \cite{DUMIC2021107606}. In this work, it was shown that using ensembles it was again possible to improve the results compared to single DRs. Moreover, for this problem the SEC method also achieved the best overall results.
 
In \cite{GilGala2019}, the authors propose a novel type of ensemble collaboration which they apply to solve the one machine problem with variable capacity. Since the considered problem is static in nature, i.e., all information about it is known in advance, the authors outline that it is possible to solve the problem instances with a selected set of DRs and then choose the best solution for each instance. The authors use a genetic algorithm to determine the best ensemble of DRs, and show that such ensembles significantly improve results over individual DRs. These ensembles are further investigated in \cite{GilGala2020}, where the authors analyse the obtained ensembles and compare ensembles created from the ATC rule and from GP generated DRs. The results show that ensembles created from automatically generated DRs perform better than those obtained from the ATC rule. Finally, in \cite{Gil-Gala2020a}, the authors investigate different methods for creating ensembles from DRs, namely an iterated greedy method, local search, and a memetic algorithm. The results show that all methods construct ensembles of good quality, with the memetic algorithm achieving the best ensembles overall.

Apart from their application in scheduling, ensembles of heuristics generated by GP have also been used for the capacitated arc routing problem \cite{Wang2019, Wang2019a}, where they were considered to evolve smaller and more interpretable rooting policies without reducing their performance.

\section{Background}
\label{sec:background}
\subsection{Unrelated machines environment}

The unrelated machines environment represents an important scheduling problem that arises in many real-world applications such as manufacturing, cloud environments, and the like \cite{Pinedo2012}. This problem consists of $n$ jobs and $m$ machines, where each job $j$ must be assigned to a particular machine $i$ \cite{Pinedo}. Each job has the following properties: 
\begin{itemize} \item processing time ($p_{ij}$) - time needed to execute job $j$ on machine $i$, \item release time ($r_j$) - time when job $j$ will be released in the system, \item due date ($d_j$) - time by which job $j$ must be executed, \item weight ($w_j$) - importance of job $j$.
\end{itemize}
The aforementioned properties are taken into account when creating a schedule for a given problem, since they have a direct impact on the quality of the schedule, depending on which criterion is optimised. In this study, the total weighted tardiness (TWT) criterion is to be minimised. TWT is defined as $TWT=\sum_{j}w_{j}\max(C_j-d_j,0)$, where $C_j$ represents the completion time of job $j$. Thus, minimising this criterion reduces the time that jobs spend executing after their due date. The problem described can be classified as $R|r_j|TWT$ using the standard notation of scheduling problems \cite{Pinedo2012}.

In addition to the above properties of the problem, another important property of the considered problem is that it is dynamic. This means that until a job is released in the system, no information about it is known about it. Moreover, it is also not known when a job will be released into the system. Therefore, at each decision point, the decision can only be made based on the jobs that have been released so far, and no information about the future of the system can be used. This feature makes DRs a suitable choice for solving such problems.

\subsection{Designing DRs with GP}
DRs usually consist of two parts, a schedule generation scheme (SGS) and a priority function (PF). The SGS is tasked with constructing the entire schedule, i.e., it determines at which point a scheduling decision should be made and which jobs and machines should be considered for that decision point. The SGS used in this study is denoted in Algorithm \ref{alg:standard-sgs} \cite{DURASEVIC2020}. This SGS performs a decision every time at least one job and one machine are available. At that moment, it determines which job should be scheduled on which machine. For this purpose, it uses a PF that assigns a numeric value to each job-machine pair. The pair that received the lowest value is selected and the corresponding job is then scheduled on the selected machine. However, the job is scheduled only scheduled if the selected machine is free. Otherwise, this job is skipped and the next one is scheduled. The entire process is repeated as long as there are jobs left.

\begin{algorithm}
	\caption{SGS used by generated DRs}
	\label{alg:standard-sgs}
	\begin{algorithmic}[1]
		\WHILE{unscheduled jobs are available}
		\STATE Wait until at least one job and machine are available
		\FOR{all available jobs $j$ and each machine $i$ in $m$}
		\STATE Get the priority $\pi_{ij}$ of scheduling $j$ on machine $i$
		\ENDFOR
		\FOR{all available jobs}
		\STATE Determine the machine with the best $\pi_{ij}$ value
		\ENDFOR
		\WHILE{jobs whose best machine is available exist}
		\STATE Determine the best priority of all such jobs
		\STATE Schedule the job with best priority
		\ENDWHILE
		\ENDWHILE
	\end{algorithmic}
\end{algorithm}

Since the PF determines the order in which jobs are scheduled, it is an important aspect of DR. Although several good PFs have been designed manually, such as those used in the ATC or COVERT DRs, it has been shown that automatically generated DRs can outperform them. For this reason, GP is often used to automatically design PFs that are appropriate for the problem under consideration. Thus, the most important part of adapting GP for such a problem is selecting the appropriate set of terminal and function nodes. The set of terminal nodes used is given in Table \ref{tbl:terminals}. The function set, on the other hand, consists of the addition, subtraction, multiplication, division (implemented as a protected division that returns 1 if the division by 0 occurs), and positive (unary operator that returns 0 if the argument is negative) operators. Both sets were selected based on previous studies \cite{DURASEVIC2016}. 

\begin{table}[]
\caption{Terminal set}
\label{tbl:terminals}
\begin{center}
\adjustbox{max width=\columnwidth}{%
\begin{tabular}{@{}c c@{}}
\toprule
\textbf{Terminal} & \textbf{Description} \\
\hline
		$pt$ & processing time of job $j$ on machine $i$ \\
		$pmin$ & minimal processing time (MPT) of job $j$ \\
		$pavg$ & average processing time of job $j$ across all machines \\
		$PAT$ & time until machine with the MPT for job $j$ becomes available \\
		$MR$ & time until machine $i$ becomes available \\
		$age$ & time which job $j$ spent in the system \\
		$dd$ & time until which job $j$ has to finish with its execution \\
		$w$ & weight of job $j$ ($w_j$) \\
		$SL$ & slack of job $j$, $-max(d_{j} - p_{ij} - t, 0)$ \\
\bottomrule
\end{tabular}}
\label{tab:terminalset}
\end{center}
\end{table}

\section{Ensemble collaboration method}
\label{sec:collab}
From the literature review, it appears that DRs in the ensemble were used to collaborate in two different ways to solve different scheduling problems. In the first way, the ensembles are used to jointly make decisions during the scheduling process \cite{Park2015, Durasevic2017, Durasevic2019}. This means that all DRs included in the ensemble are evaluated at each decision point and then a single decision is made based on their individual decisions. To combine these decisions, different ensemble combination methods, such as linear combination or majority voting, can be used. On the other hand, in the second type of ensembles \cite{GilGala2019, GilGala2020, Gil-Gala2020a}, each rule is applied individually to the problem instance and in the end the best solution among the rules included in the ensemble is selected as the final solution. However, such a method is only applicable in the case of static scheduling conditions, since all information about the system must be known in advance.

From the preceding description, it is clear that both types of ensembles are motivated by the fact that no single DR can perform well on all problems, but the way the rules of the ensembles work together is different. While the first type of ensembles has been used under both static and dynamic conditions, the second type of ensembles has only been used for static problems due to their unique characteristics. However, the concept of executing rules in parallel and selecting the best result would also be interesting to consider in dynamic environments as well.

In dynamic scheduling environments, the DR does not have the information about which jobs will arrive in the future. However, when jobs are released, they wait to be scheduled until a machine becomes free. Therefore, it is to be expected that as the system runs, a certain number of jobs will accumulate in the queue, waiting for the moment when they are selected for scheduling. Traditional DRs operate in such a way that, based on the current status, they schedule a job only for the available machine and wait again when a machine becomes free. Therefore, an ensemble could be defined so that all DRs in it suggest which job should be scheduled next. However, instead of combining these decisions into one, a method can be used to determine how good a decision each rule made. The simplest method would be to determine how each decision affects the optimised criterion and select the one that results in the smallest increase in the optimised criterion. However, using only a single job to determine how good a decision is can lead to poor decisions. Therefore, each rule can be used to simulate not only the scheduling of the next job, but rather all currently released jobs under the assumption that no new jobs will be released in the future. In this way, it would be possible to get a more accurate approximation of how the next decision might affect the quality of the schedule in the future and how well each DR performs under current system conditions. It must be emphasised that no job is actually scheduled during this process, but rather the behaviour of each DR is merely simulated with the current system condition. Only when the DR with the best decision has been determined on the basis of the simulation is this decision implemented in the real schedule.

%However, at each decision point the DR could be used to build the entire schedule as if no future jobs would be released and the objective value of the constructed schedule could be used as an approximation of how a good decision was performed at the current scheduling moment. Therefore, it would be possible to execute several DRs independently to simulate how good they would schedule the currently available jobs. Based on those simulations the quality of the decision performed by each DR can be assessed, and the appropriate scheduling decision can be performed.

Based on the previous description, we propose an ensemble collaboration method that uses multiple DRs in parallel to build the schedule. Algorithm \ref{alg:drens} gives an overview of the proposed ensemble collaboration method, which is similar to the standard SGS. The main difference is that at each decision point, each DR in the ensemble is used to create the schedule considering only the currently released jobs. It should be noted that the DRs here only simulate the schedule and that no jobs are actually scheduled in the real system. After running the simulations of all the rules, the rule that gave the best objective value in the simulation is selected. This rule is then used to make the scheduling decision at the current time. It should be emphasised that the rule schedules only a single job, and then moves to the next decision point where the entire procedure is repeated. Therefore, any new jobs released between these two decision points are also considered in the next decision point, which means that the procedure is suitable for dynamic environments.

%In addition, such a methodology also allows that at each decision point a different DR performs the decision.

\begin{algorithm}
	\caption{SGS for the ensemble collaboration method}
	\label{alg:drens}
	\begin{algorithmic}[1]
		\WHILE{unscheduled jobs are available}
		\STATE Wait until at least one job and machine are available
		\FOR{each rule $k$ in the ensemble $E$}
		\STATE Build the schedule with only the released jobs at the current moment
		\STATE Store the quality of the simulated schedule in $F[k]$
		\ENDFOR
		\STATE Determine the rule $k_{best}$ which has the lowest value objective value of the simulated schedule
		\STATE Apply rule $k_{best}$ to determine the next scheduling decision 
		\ENDWHILE
	\end{algorithmic}
\end{algorithm}

Although not explicitly stated in the algorithm, there is a parameter that can be used to change the behaviour of the proposed method. As mentioned in the previous descriptions, the approximation how good a DR is at the current time can be done by considering a different number of currently released jobs. Although this parameter can be set to any value, in this study we will examine only the two extreme values, namely by scheduling only the next job and evaluating its impact on the schedule, and scheduling all released jobs and evaluating the performance of such a schedule. The first method is referred to as EDR-S, while the second method is referred to as EDR-M.

Based on the previous descriptions, it can be seen that the proposed ensemble collaboration method has certain similarities with the rollout heuristic \cite{Durasevi2020a}. At each decision point, the rollout heuristic considers all possible decisions, but to evaluate how good each of these decisions is, it uses a DR, to approximate the rest of the schedule. For this reason, the method is only applicable in static scheduling environments. The proposed ensemble collaboration method can be considered as a particular way of combining ensembles and the rollout heuristic, since it uses multiple DRs from ensembles at once to approximate the rest of the schedule similarly to the rollout heuristic. By combining these two concepts, it was possible to develop a method that is now applicable to dynamic environments. 

An open question is still how to construct the ensembles. The advantage of the proposed approach is that most of the methods proposed in previous studies can be used to construct such ensembles. Since in this study we mainly focus on demonstrating the effectiveness of using the proposed ensemble collaboration method, the SEC method is used because of its simplicity, but also because of its ability to construct better ensembles than other more sophisticated methods \cite{Durasevic2017}. SEC is applied using the random construction method. This method construct $p$ ensembles, where $p$ is a user-specified parameter, where each of the ensembles is constructed by randomly selecting $e$ DRs from the pool of available rules \cite{Durasevic2019}. Here $e$ represents the size of the ensemble to be constructed and is also a user-specified parameter.

\section{Experimental study}
\label{sec:exp}
\subsection{Setup}

To develop rules, create ensembles, and evaluate their efficiency, a problem set of 180 instances was used. These instances contain between 12 and 100 jobs, and between 3 and 10 machines. In addition, the due dates for each instance were generated with different parameters, resulting in instances with different levels of difficulty. The instances are divided into 3 independent sets, the test set, the validation set and the training set. The training set is used by GP to develop new DRs. More specifically, GP is executed 50 times, and the best individual from each execution is stored. These DRs are used to create ensembles with the SEC method using the validation set. Finally, the test set is used to measure the quality of the created DRs and ensembles. All values mentioned in this section were obtained using this set.

For the development of DRs, the steady-state algorithm GP with a 3-tournament selection is used. The parameters of the algorithm are listed in Table \ref{tbl:auto-gpParam}. Since multiple operators are used for crossover and mutation, one is randomly selected from the set of defined operators each time such an operator needs to be applied. After 50 DRs have been developed, the SEC method uses them as building blocks for constructing the ensembles. The parameters that must be specified for the SEC method are the number of DRs constructed per execution and the size of the constructed ensembles. For the number of constructed ensembles, the values 100, 500, 1000, 5000, 10000, and 20000 are used. For the ensemble size, the initial tests are performed with sizes 3, 5, and 7. The SEC method is executed 30 times to obtain statistically significant results. From each execution, the best ensemble obtained on the validation set is saved and evaluated on the test set. The tables then show the minimum, median and maximum values of these 30 runs. 

\begin{table}[]
	\centering
	\caption{Parameters for GP}
	\label{tbl:auto-gpParam}
	\begin{tabular}{@{}p{3cm}p{5cm}@{}}
		\toprule
		Parameter name            & Parameter value\\ \midrule
		Population size       & 1000 individuals                                                                       \\
		Termination criterion & 80 000 function evaluations                                                            \\
		Initialisation        & Ramped half-and-half                                                        \\
		Maximum tree depth    & 5                                                                           \\
		Crossover operators   & Subtree, uniform, context-preserving, size-fair                             \\
		Mutation operators    & Subtree, hoist, node complement, node replacement, permutation, shrink \\
		Mutation probability  & 0.3 \\
		\bottomrule
	\end{tabular}
\end{table}

To test whether significant differences exist between the different parameter values, the Kruskal-Wallis test is used. The Bonferroni correction method is in post-hoc analysis. In addition, the Mann-Whitney test is used for individual pairwise comparisons. Results are considered significantly different if a p-value of less than 0.05 is obtained.

\subsection{Results}

Table \ref{tbl:ressingle} shows the results for three selected ensemble sizes using the EDR-S method. The results show that there is little difference between ensembles created using the SEC method with different numbers of constructed ensembles. This shows that even a smaller number of ensembles is sufficient to find those that perform better than individual DRs. The best overall results were obtained with the construction of 1000 ensembles of size 5. The statistical test showed that, in general, there was no significant difference between the results obtained using the SEC method with different numbers of constructed ensembles. However, in all cases significantly better results were obtained compared to individual DRs. Thus, it can be concluded that the SEC method can produce ensembles that are better than individual rules, even with a small number of samples.

\begin{table*}[]
    \centering
    \caption{Results for the EDR-S collaboration method}
    \label{tbl:ressingle}
	\begin{tabular}{@{}llllllllllll@{}}
		\toprule
		Method         & \multicolumn{11}{c}{Ensemble size}                                          \\ \midrule
		& \multicolumn{3}{c}{3} &  & \multicolumn{3}{c}{5} &  & \multicolumn{3}{c}{7} \\ \cmidrule(lr){2-4} \cmidrule(lr){6-8} \cmidrule(l){10-12} 
		& Min   & Med   & Max   &  & Min   & Med   & Max   &  & Min   & Med   & Max   \\ \midrule
		Individual DRs & 15.23 & 15.94 & 17.59 &  & 15.23 & 15.94 & 17.59 &  & 15.23 & 15.94 & 17.59 \\
		SEC-100            & 14.97 & 15.39 & 15.96 &  & \textbf{14.96} & 15.39 & 16.17 &  & 15.03 & 15.52 & 15.95 \\
		SEC-500            & 15.00 & 15.40 & 16.04 &  & 15.08 & 15.49 & 15.80 &  & 14.99 & \textbf{15.42} & 16.04 \\
		SEC-1000           & 14.86 & 15.34 & \textbf{15.88} &  & 15.01 & \textbf{15.29} & \textbf{15.63} &  & \textbf{14.79} & 15.44 & \textbf{15.88} \\
		SEC-5000           & 14.93 & 15.35 & 15.89 &  & 15.00 & 15.49 & 16.09 &  & 15.07 & 15.43 & 16.00 \\
		SEC-10000          & \textbf{14.80} & \textbf{15.27} & 15.93 &  & 15.01 & 15.31 & 15.82 &  & 15.07 & 15.44 & 15.91 \\
		SEC-20000          & 14.96 & 15.41 & 15.91 &  & 14.97 & 15.38 & 15.84 &  & 15.14 & \textbf{15.42} & 16.01 \\ \bottomrule
	\end{tabular}
\end{table*}

Table \ref{tbl:resmultiple} shows the results obtained with the EDR-M method. The table shows that the results obtained with the different number of constructed ensembles for the SEC method are quite similar. As in the previous method, the best overall median value is again obtained for the ensemble of size five, but this time constructed with SEC-500. However, the statistical test showed that for all three ensemble sizes there is no significant difference between the results obtained for the different SEC parameter values. Nevertheless, in all cases, the SEC method produced ensembles that were significantly better than the individual DRs.

\begin{table*}[]
    \centering
    \caption{Results for the EDR-M collaboration method}
    \label{tbl:resmultiple}
	\begin{tabular}{@{}llllllllllll@{}}
		\toprule
		Method         & \multicolumn{11}{c}{Ensemble size}                                           \\ \midrule
		& \multicolumn{3}{c}{3} &  & \multicolumn{3}{c}{5} &  & \multicolumn{3}{c}{7}  \\ \cmidrule(lr){2-4} \cmidrule(lr){6-8} \cmidrule(l){10-12} 
		& Min   & Med   & Max   &  & Min   & Med   & Max   &  & Min   & Med    & Max   \\ \midrule
		Individual DRs & 15.23 & 15.94 & 17.59 &  & 15.23 & 15.94 & 17.59 &  & 15.23 & 15.94  & 17.59 \\
		SEC-100            & 14.88 & 15.38 & 15.78 &  & 14.72 & 15.21 & 15.96 &  & 14.84 & 15.22  & 15.62 \\
		SEC-500            & 14.72 & 15.33 & 15.81 &  & 14.72 & \textbf{15.04} & 15.71 &  & \textbf{14.66} & 15.17  & 16.03 \\
		SEC-1000           & \textbf{14.69} & 15.32 & 15.70 &  & 14.80 & 15.23 & 15.85 &  & 14.85 & \textbf{15.10}  & 15.64 \\
		SEC-5000           & 14.78 & 15.36 & \textbf{15.61} &  & 14.72 & 15.18 & 15.65 &  & 14.84 & 15.18  & \textbf{15.53} \\
		SEC-10000          & 14.78 & 15.25 & \textbf{15.61} &  & \textbf{14.71} & 15.11 & 15.75 &  & 14.81 & 15.17  & 15.75 \\
		SEC-20000          & 14.78 & \textbf{15.17} & \textbf{15.61} &  & 14.75 & 15.15 & \textbf{15.67} &  & 14.88 & \textbf{15.10} & 15.58 \\ \bottomrule
	\end{tabular}
\end{table*} 

\begin{figure}
	\centering
	\begin{subfigure}[b]{\columnwidth}
		\includegraphics[width=\textwidth]{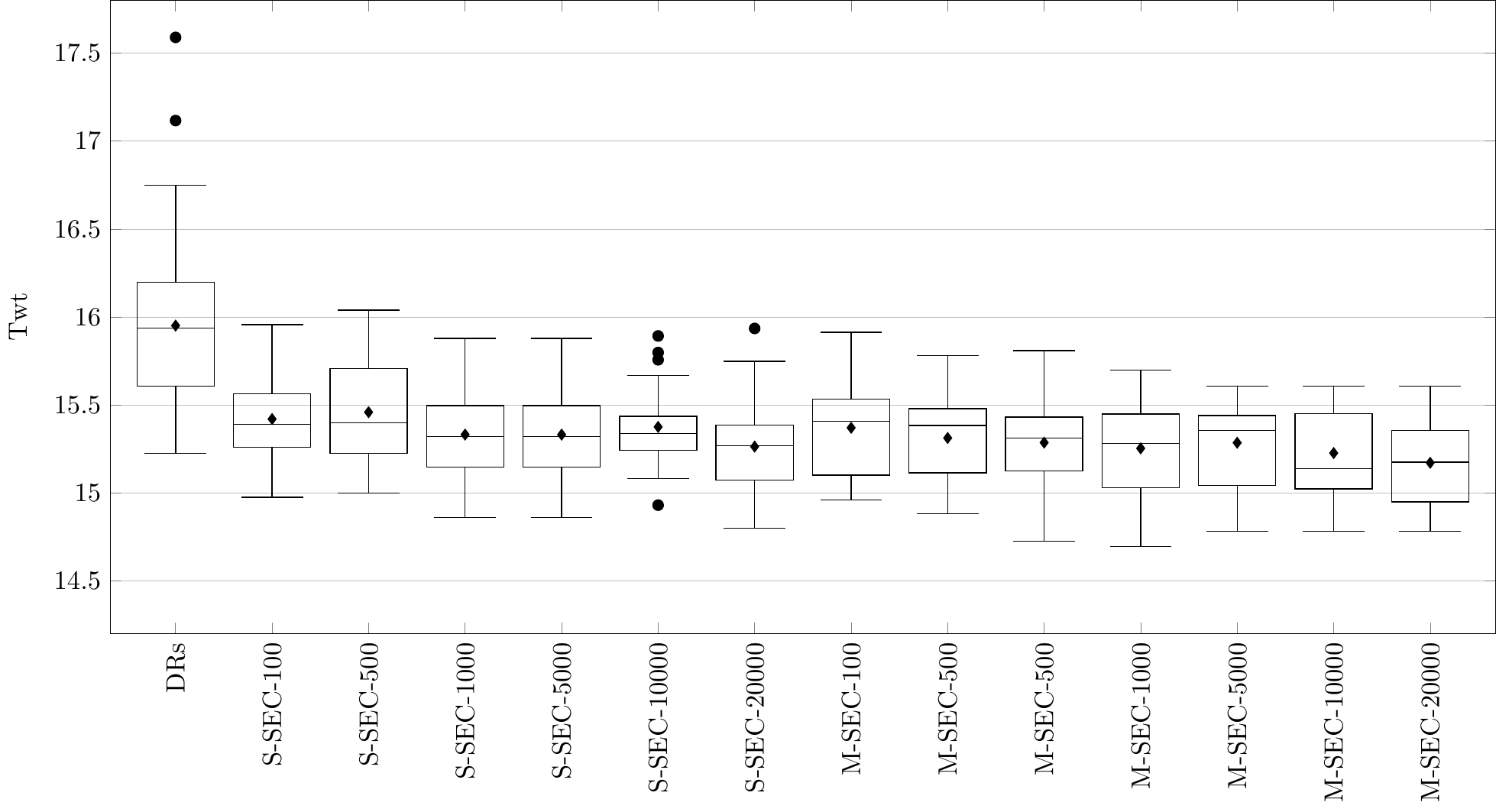}
		\caption{Results for ensemble size 3}
		\label{fig:SEC3}
	\end{subfigure}
	\begin{subfigure}[b]{\columnwidth}
		\includegraphics[width=\textwidth]{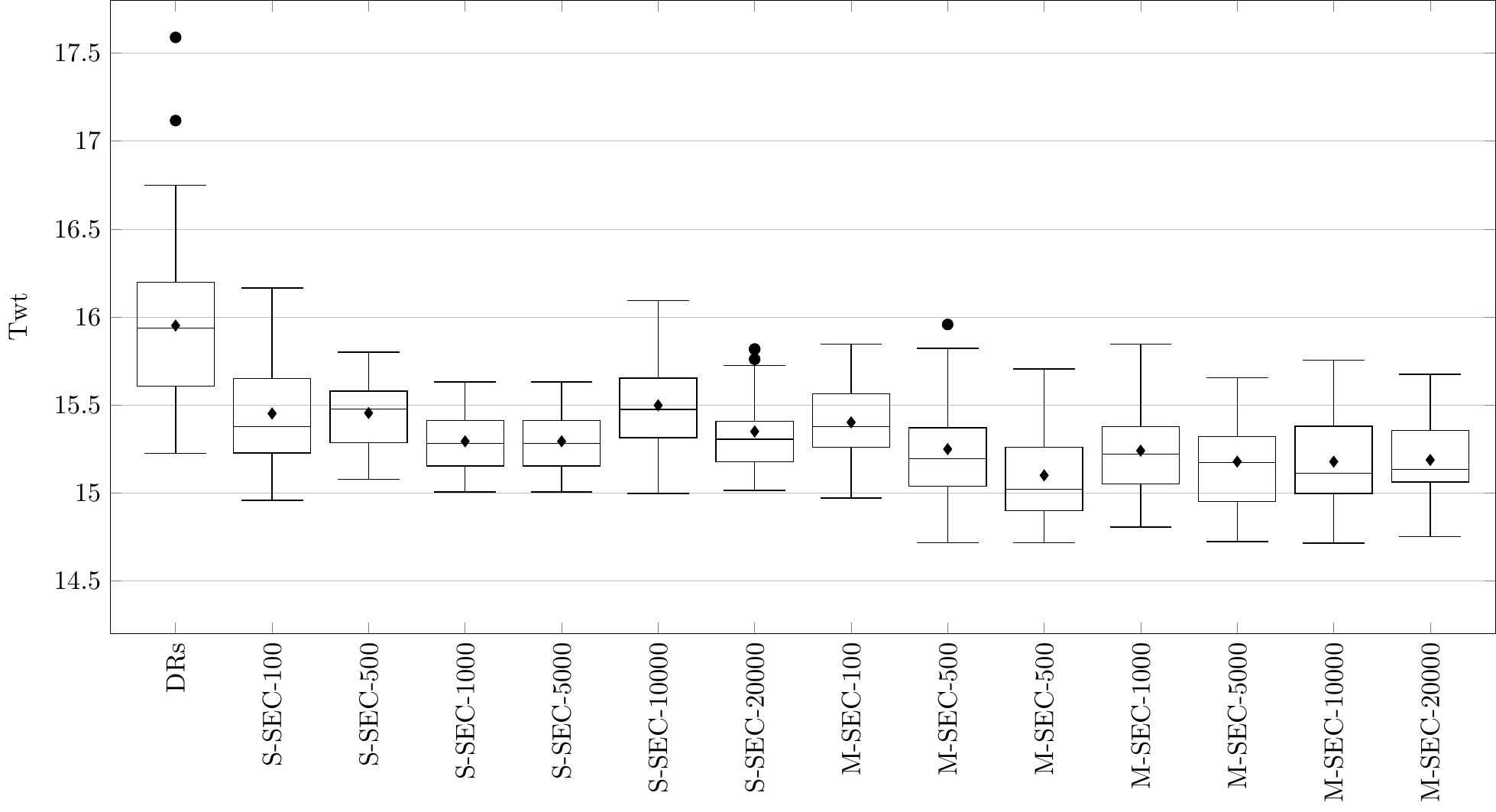}
		\caption{Results for ensemble size 5}
		\label{fig:SEC5}
	\end{subfigure}
	\begin{subfigure}[b]{\columnwidth}
		\includegraphics[width=\textwidth]{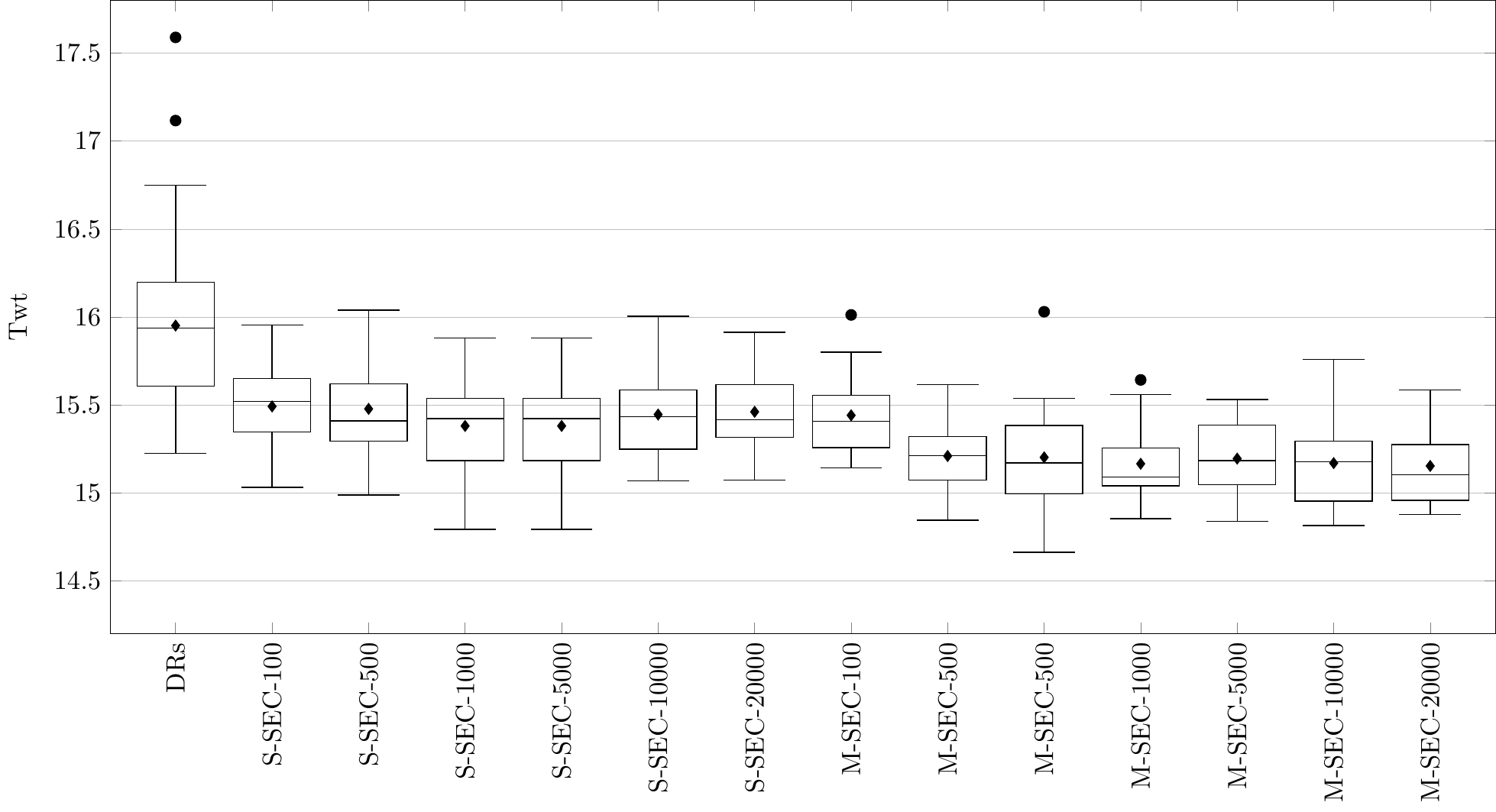}
		\caption{Results for ensemble size 7}
		\label{fig:SEC7}
	\end{subfigure}
	\caption{Boxplot representation of the results}
	\label{fig:box}
\end{figure}

To get a better idea of how the two tested methods compare, Figure \ref{fig:box} shows the results for them grouped by the different ensemble sizes. In this figure, S- SEC denotes the variant of SEC that uses the EDR-S method to evaluate ensembles, and M- SEC denotes the variant that uses the EDR-M method. For ensembles of size 3, it can be seen that there is not much difference between the methods for the number of ensembles constructed. However, the situation is different for the two larger ensembles. In these cases, we can see a significant difference between the results of the two methods. For ensemble size 5, EDR-M achieves significantly better results for all experiments except when 1000 ensembles were constructed. For ensemble size 7, EDR-M achieves significantly better results in all cases.

Based on the previous observations, the following conclusions can be drawn about the two tested. In both cases, the number of sampled ensembles in the SEC method has no significant influence on the results, which means that in both cases a smaller number of ensembles can be sampled, which is advantageous from the point of view of the execution time of the method. As for the ensemble sizes, it can be observed that there are no significant differences between the two methods at the smallest ensemble size. Thus, it seems that when the number of rules in the ensemble is small, the choice of which rule to select from the ensemble can be made even when considering a single job. However, this is not the case for the two larger ensemble sizes, as the results obtained with EDR-M are always significantly better than those obtained with EDR-S. This seems to indicate that it is more difficult to make the right decision about which DR to select from the ensemble once there are more rules in the ensemble. This makes sense because, again, it is difficult to evaluate multiple rules based on a single decision they make.

The obtained results are comparable or even better in several cases when compared to ensembles proposed in \cite{Durasevic2017a} and \cite{Durasevic2019}. However, we leave further comparisons between the different ensemble collaboration methods and deeper analyses for future work, as the main objective of this paper was to focus on the proposed collaboration method. 

\section{Further analysis} 

\subsection{Influence of the ensemble sizes}

To investigate the influence of ensemble size, ensembles with sizes between 2 and 20 DRs are tested. Using the SEC method, 500 ensembles are constructed for the EDR-M method and 1000 ensembles are constructed for the EDR-S method, as the methods produced their best median results at these values. 

The results for different ensemble sizes are shown in Table \ref{tbl:ensemblesizes}. Size 1 in the table denotes the results obtained with each DR. The results show that the constructed ensembles always achieve a better median value compared to the individual DRs. However, the statistical tests show that the difference is not always significant. For the EDR-M method, the constructed ensembles always perform better than the individual DRs. On the other hand, the EDR-S method achieves significantly better results than the individual DRs for all ensemble sizes except for sizes 14, 17, 18, 19, and 20. The best median value is obtained for ensemble size 5 in both cases. This shows that already with smaller ensemble sizes quite good results can be obtained and that the use of larger ensemble sizes does not bring any real advantage.

\begin{table}[]
    \centering
    \caption{Results for different ensemble sizes}
    \label{tbl:ensemblesizes}
	\begin{tabular}{@{}lccccccc@{}}
		\toprule
		& \multicolumn{3}{c}{EDR-S} &  & \multicolumn{3}{c}{EDR-M} \\ \cmidrule(lr){2-4} \cmidrule(l){6-8} 
		& Min     & Med     & Max    &  & Min      & Med     & Max     \\ \midrule
		1  & 15.23   & 15.94   & 17.59  &  & 15.23    & 15.94   & 17.59   \\
		2  & 14.95   & 15.58   & 15.93  &  & 14.91    & 15.43   & 15.71   \\
		3  & 14.86   & 15.34   & 15.88  &  & 14.72    & 15.33   & 15.81   \\
		4  & 14.99   & 15.45   & 15.83  &  & 14.84    & 15.27   & 15.64   \\
		5  & 15.01   & \textbf{15.29}   & 15.63  &  & 14.72    & \textbf{15.04}   & 15.71   \\
		6  & 14.81   & 15.36   & 15.92  &  & 14.84    & 15.13   & 16.03   \\
		7  & 14.79   & 15.44   & 15.88  &  & 14.66    & 15.17   & 16.03   \\
		8  & 15.03   & 15.42   & 15.91  &  & 14.89    & 15.18   & 15.72   \\
		9  & 15.12   & 15.46   & 15.96  &  & 14.86    & 15.13   & 15.68   \\
		10 & 15.16   & 15.46   & 16.19  &  & 14.94    & 15.15   & 15.74   \\
		11 & 15.06   & 15.54   & 16.19  &  & 14.84    & 15.26   & 15.58   \\
		12 & 15.02   & 15.58   & 16.16  &  & 14.82    & 15.19   & 15.37   \\
		13 & 15.20   & 15.51   & 16.07  &  & 14.85    & 15.20   & 15.49   \\
		14 & 15.26   & 15.70   & 16.12  &  & 14.97    & 15.24   & 15.51   \\
		15 & 14.99   & 15.63   & 16.15  &  & 14.90    & 15.31   & 15.57   \\
		16 & 15.18   & 15.62   & 16.06  &  & 14.77    & 15.23   & 15.55   \\
		17 & 15.21   & 15.81   & 16.72  &  & 14.87    & 15.21   & 15.67   \\
		18 & 15.17   & 15.73   & 16.20  &  & 14.94    & 15.17   & 15.48   \\
		19 & 15.17   & 15.70   & 16.29  &  & 14.86    & 15.22   & 15.56   \\
		20 & 15.34   & 15.76   & 16.17  &  & 14.91    & 15.26   & 15.60   \\ \bottomrule
	\end{tabular}
\end{table}

\begin{figure} 
	\centering
	\begin{subfigure}[b]{\columnwidth}
		\includegraphics[width=\textwidth]{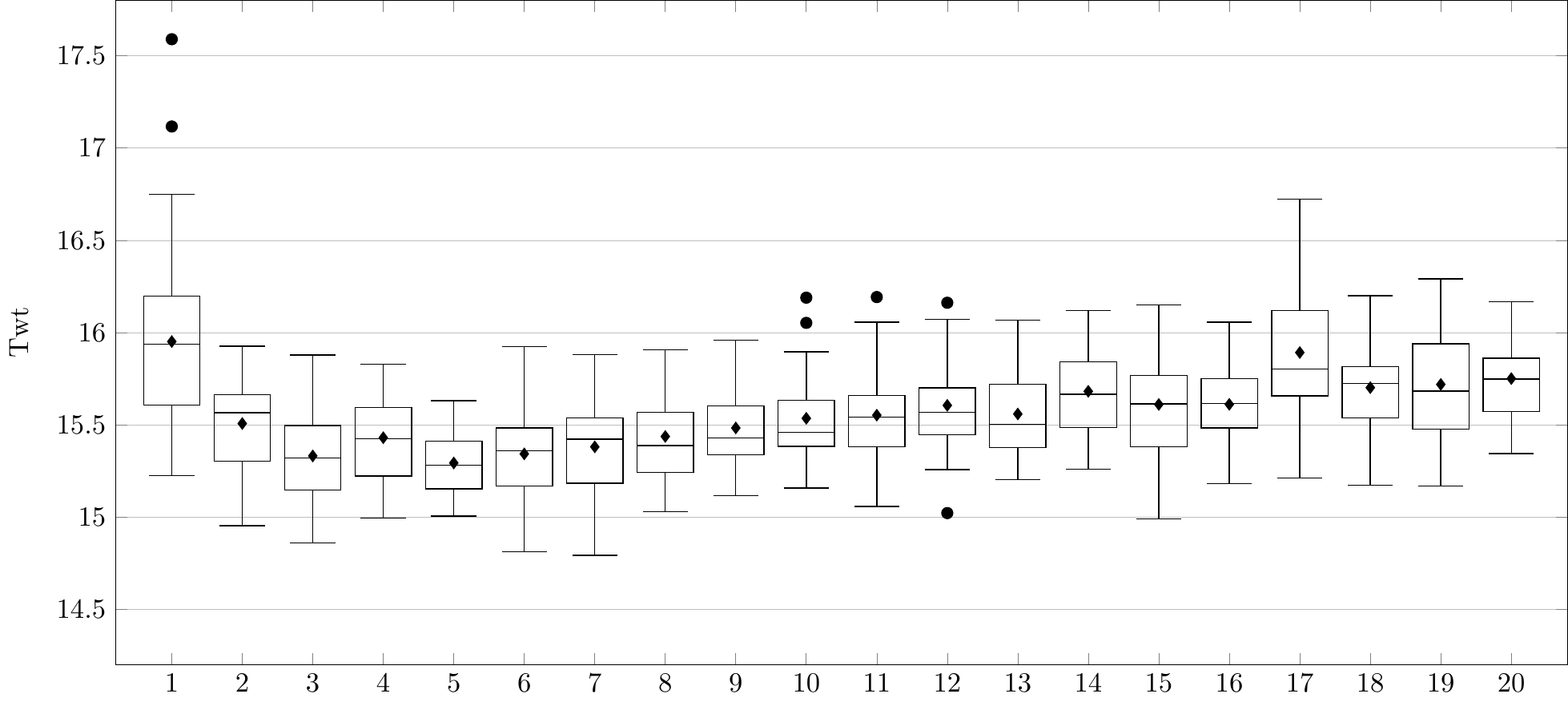}
		\caption{Results for the EDR-S method}
		\label{fig:collv1}
	\end{subfigure}
	\begin{subfigure}[b]{\columnwidth}
		\includegraphics[width=\textwidth]{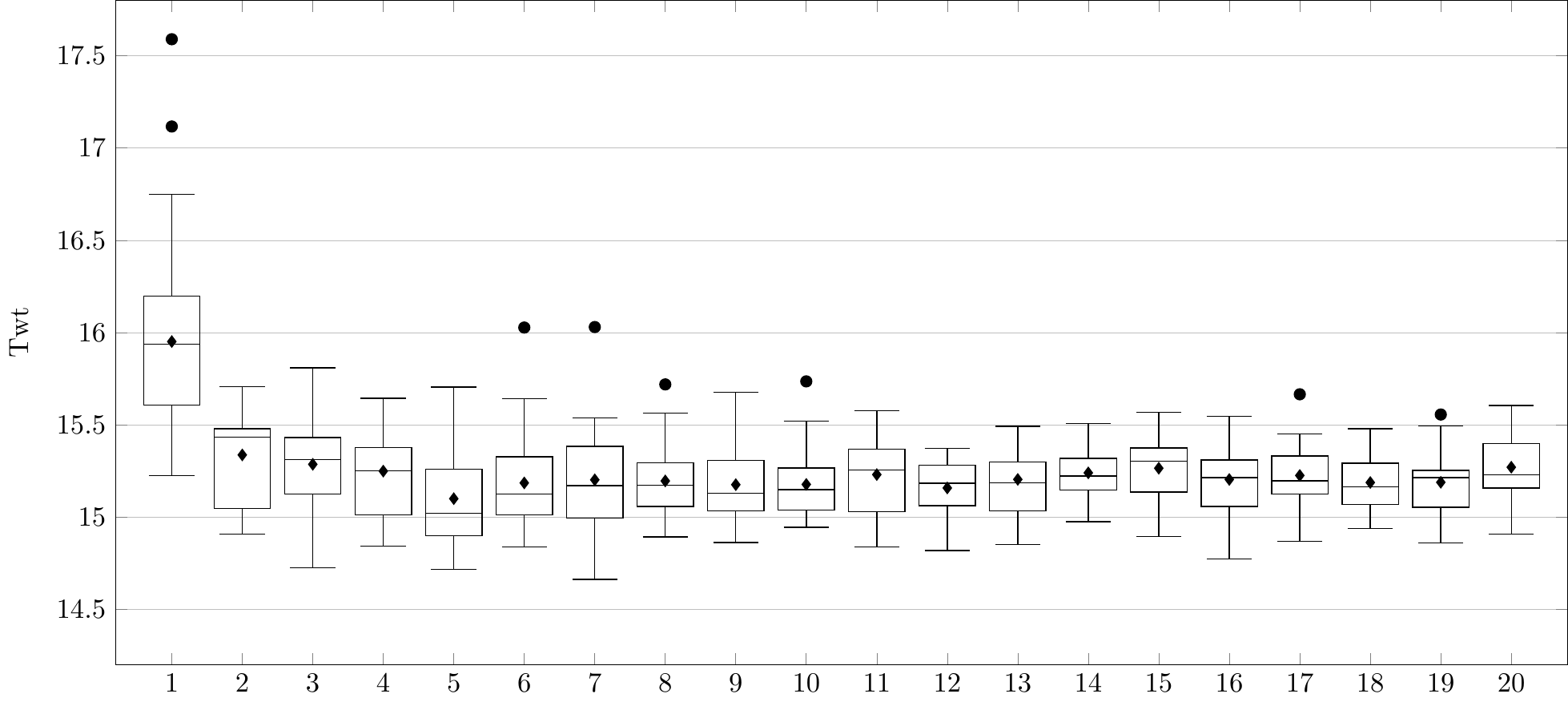}
		\caption{Results for the EDR-M method}
		\label{fig:collv1}
	\end{subfigure}
	\caption{Results for different ensembles sizes}
	\label{fig:boxsize}
\end{figure}

To better illustrate the differences between the different ensemble sizes, Figure \ref{fig:boxsize} shows the results in the form of boxplots. For the EDR-S method, it is interesting to see how the results slowly deteriorate as the ensemble size is increased beyond 5. Thus, in general, this method does not seem to work well for larger ensemble sizes. This is because the decision of which DR to apply is based on only a single job. This leads to poor approximations and there is a greater chance that a suboptimal decision will be made. For example, a DR might schedule a job on a machine where it will run for a very long time but meet its due date. Although this may seem like a good choice from the perspective of that job, it may have a negative impact in the long run because the machine is blocked longer and some other jobs may be late. However, by considering only a signle job, it is not possible to detect and prevent such a situation.

For the EDR-M method, it can also be seen that the results improve up to size 5, after which they slowly deteriorate. However, the differences are not as significant as for the EDR-S method. This is a consequence of the fact that in this method the approximation is based on all released jobs, which gives a better idea of how the decision of each DR might affect the immediate future. Although this method is more stable with larger ensembles, we again find that the results deteriorate slightly. Thus, it appears that larger ensembles introduce more noise into the planning process, i.e., with more choices, it seems more difficult for the method to select the best one.

\subsection{DR occurrence in ensembles}
\begin{figure}
	\centering
	\begin{subfigure}[b]{\columnwidth}
		\includegraphics[width=\textwidth]{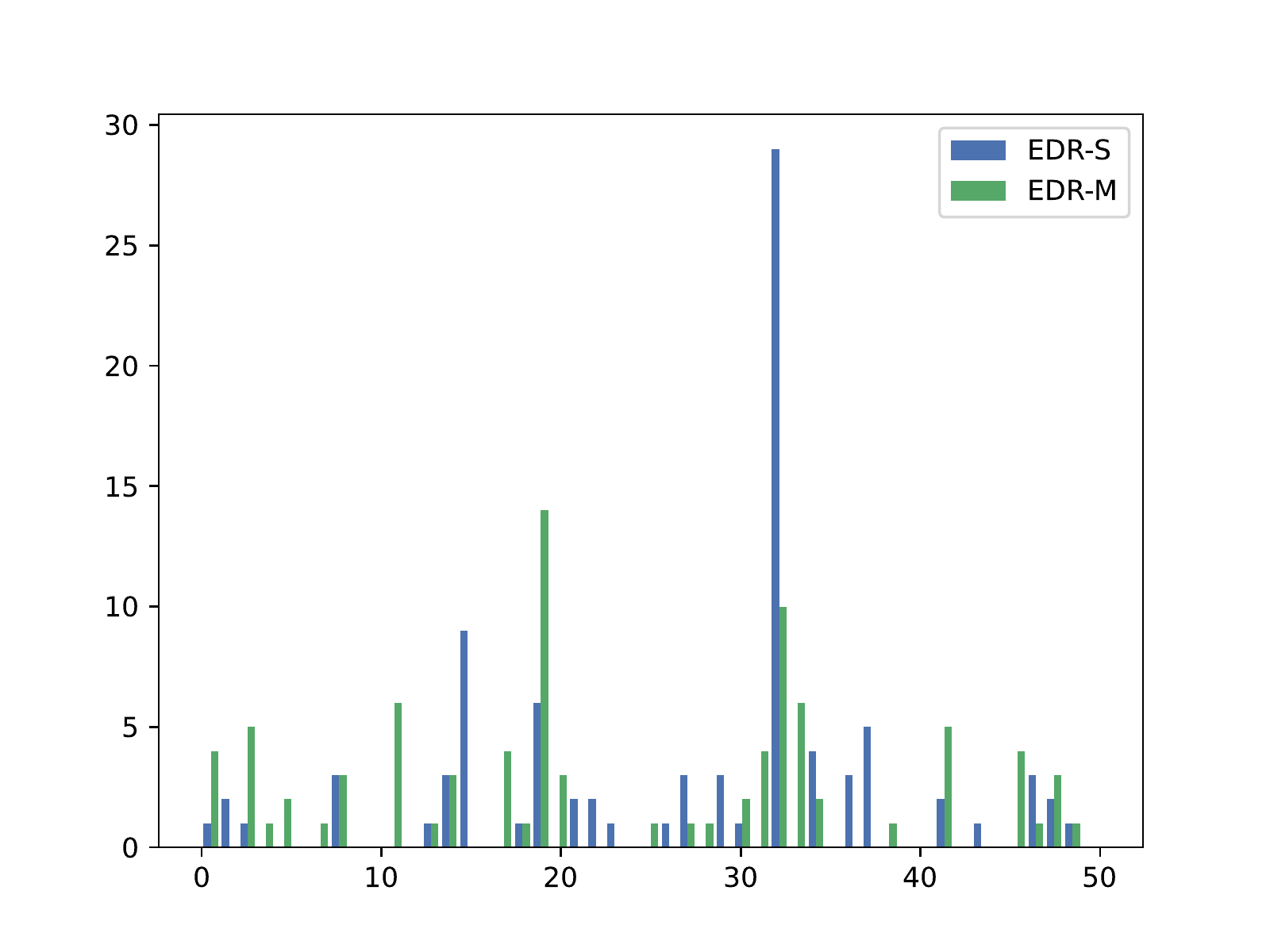}
		\caption{Results for ensemble size 3}
		\label{fig:SEC3}
	\end{subfigure}
	\begin{subfigure}[b]{\columnwidth}
		\includegraphics[width=\textwidth]{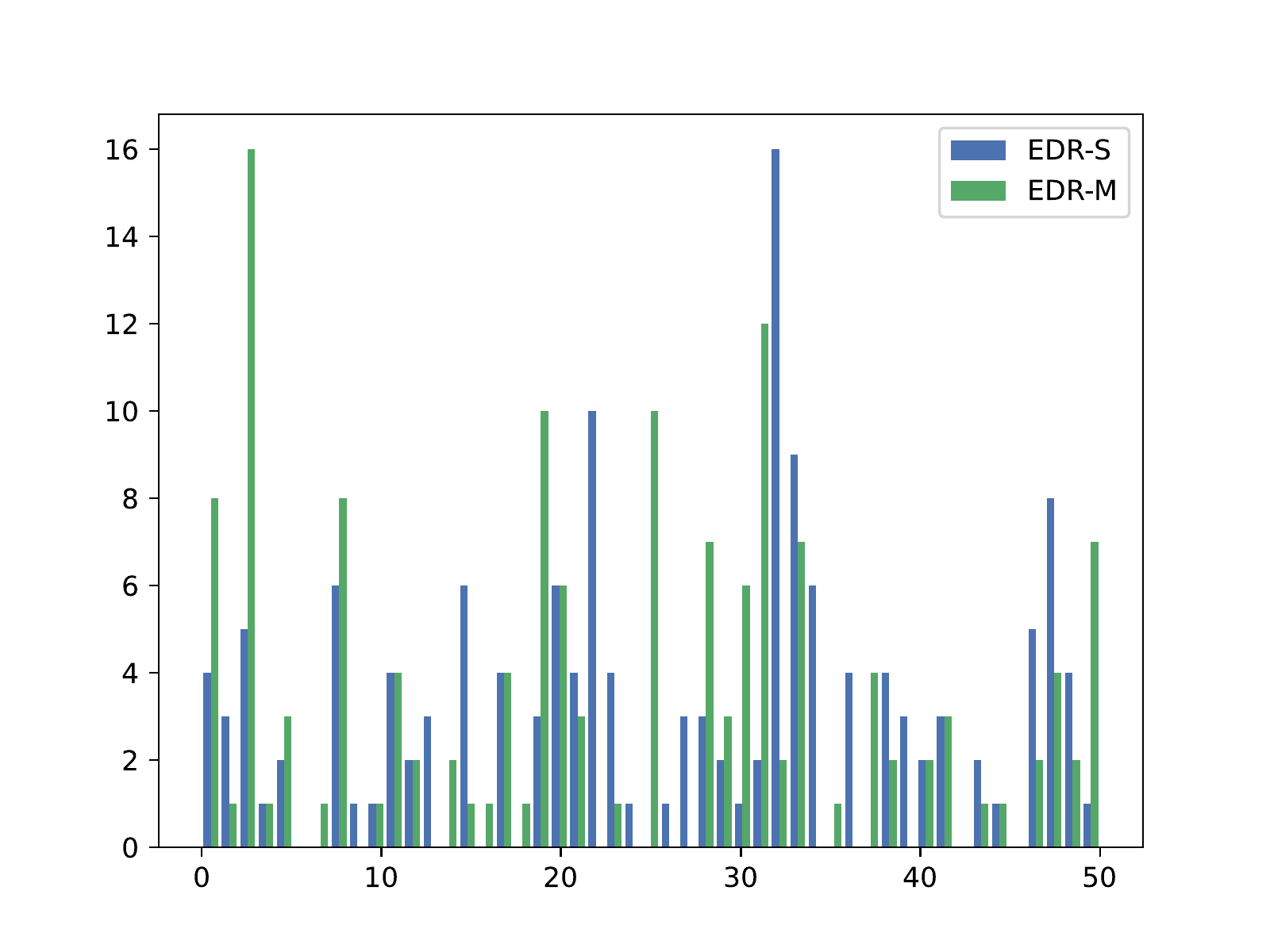}
		\caption{Results for ensemble size 5}
		\label{fig:SEC5}
	\end{subfigure}
	\begin{subfigure}[b]{\columnwidth}
		\includegraphics[width=\textwidth]{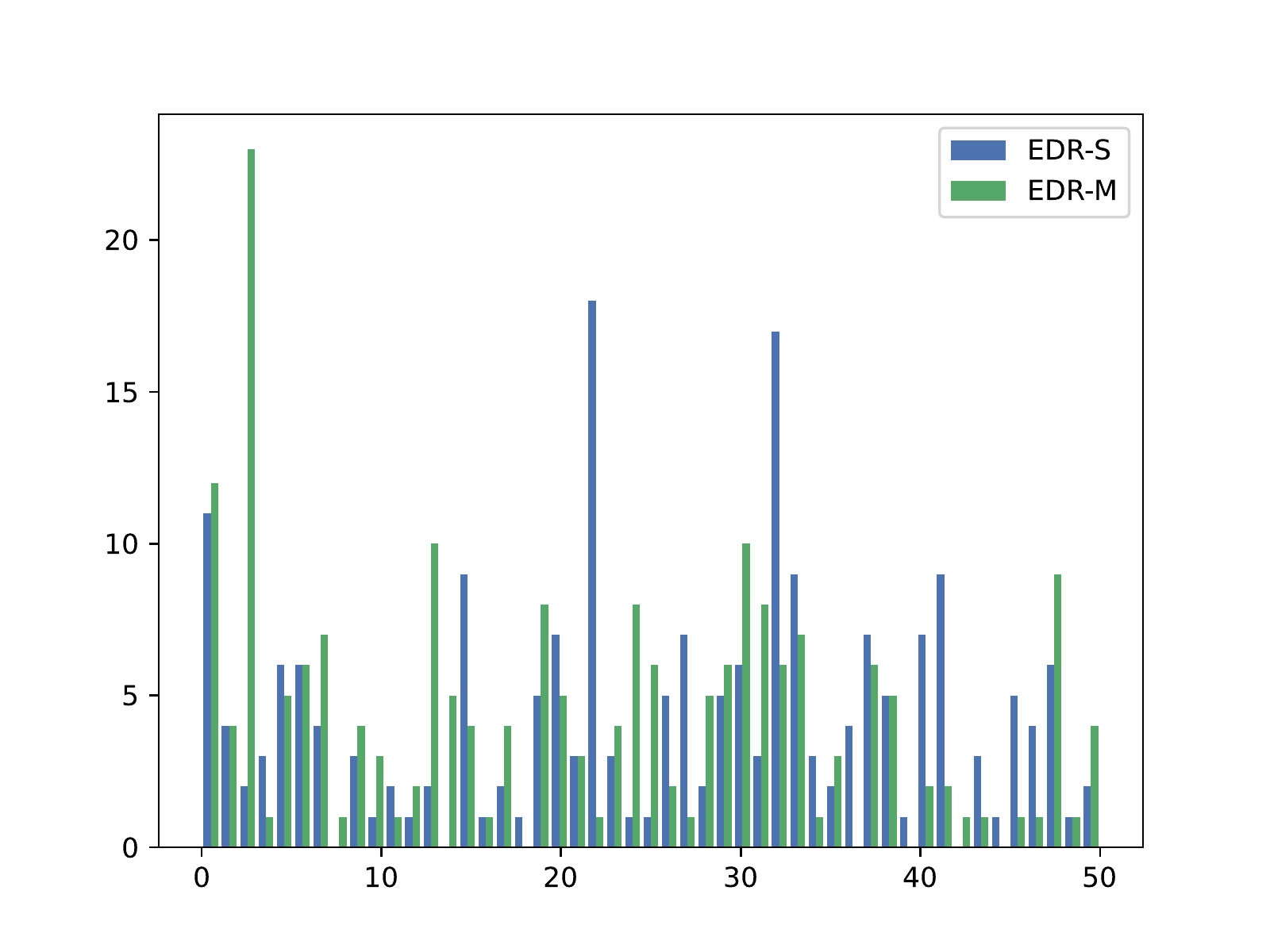}
		\caption{Results for ensemble size 7}
		\label{fig:SEC7}
	\end{subfigure}
	\caption{Frequency of occurrence of DRs in the ensembles}
	\label{fig:histo}
\end{figure}

Figure \ref{fig:histo} shows the number of occurrences for each DR in the 30 generated ensembles using S- SEC -1000 and M- SEC -500. The figures show the frequency for all three tested ensemble sizes. The figure shows that the methods clearly have a preference for selecting certain DRs. This is most evident for ensemble size 3. In this case, both methods are more biased towards selecting specific DRs. Even more, it can be seen that the EDR-S method even selects a specific DR in almost all ensembles. As the ensemble size increases, the tendency to select certain DRs decreases, but it can still be seen that the method favours certain DRs more than others. 

It is also interesting to observe that in many cases both ensemble collaboration methods select the same DRs more frequently. The same is also true for certain DRs that are rarely selected by both methods. Thus, it appears that certain DRs are inherently less likely to be part of an ensemble. Such an analysis could be used to reduce the set of eligible DRs used to form the ensembles, and thus also reduce the search space. In additional analyses, it would also be interesting to investigate whether the rules that are more or less suitable for constructing ensembles have certain properties that could be used to detect such rules.

\subsection{Ensemble execution time}

Table \ref{tbl:exec} presents the average execution time of each DR and the ensemble on the entire test set. As expected, the execution time of the ensemble is higher than that of the individual DRs. For the EDR-S method, it can be observed that the increase in execution time is not significant. This is due to the fact that each rule makes only a single decision, resulting in only a small overhead. It is interesting to observe that the increase in execution time is linearly dependent on the number of DRs in the ensemble. However, the EDR-M method incurs a much larger overhead since it must schedule multiple orders at each decision point. However, it can still be seen that all 60 instances are solved in less than three seconds for all ensemble sizes. This shows that the method is applicable in dynamic environments despite the additional computational overhead. The increase in execution time is again linear and depends on the number of DRs in the ensemble.

\begin{table}[]
    \caption{Average execution time of DRs and ensembles on the test set}
    \label{tbl:exec}
\begin{tabular}{@{}lccc@{}}
\toprule
Method & \multicolumn{3}{c}{Ensemble size} \\ \midrule
       & 3         & 5         & 7         \\ \midrule
DRs    & & 0.273 &         \\
EDR-S  & 0.352     & 0.491     & 0.633     \\
EDR-M  & 1.197     & 2.192     & 2.991     \\ \bottomrule
\end{tabular}
\end{table}

\section{Conclusion}
\label{sec:conc}
In this paper, we propose a novel ensemble collaboration method for solving dynamic scheduling problems. This method is inspired by existing ensemble learning methods used for static scheduling environments and the rollout heuristic. Experiments show that the proposed method can achieve superior results compared to individual DRs, even when using a simple method based only on random selection for ensemble construction. This shows that it is easy to find combinations of DRs that work well for the considered ensemble method, but also that such ensembles can be constructed in a small amount of time. Additional analysis of different ensemble sizes has shown that the EDR-M variant of the method is more resilient and more consistently produces good results than the EDR-S variant.

Several directions are planned to extend this work. First, different methods for creating ensembles will be investigated to determine if they can increase the quality of the results or at least create ensembles of similar quality but in less time. Second, a more in-depth analysis of the method will be conducted. This includes an analysis of the frequency with which DRs appear in the ensembles, as well as a deeper analysis of the parameter that determines how many jobs are scheduled when creating the simulated schedule and its impact on runtime. Finally, a detailed comparison between the proposed and existing ensemble methods is also performed to analyse and outline the strengths and weaknesses of both approaches.

% \begin{acks}
% This work has been supported in part by Croatian Science Foundation under the project IP-2019-04-4333 and by the Spanish Government under research projects PID2019-106263RB-I00.
% \end{acks}

%\pagebreak

\bibliographystyle{ACM-Reference-Format}
\bibliography{sample-base}


\begin{thebibliography}{42}

%%% the Web address.

\ifx \showCODEN    \undefined \def \showCODEN     #1{\unskip}     \fi
\ifx \showDOI      \undefined \def \showDOI       #1{#1}\fi
\ifx \showISBNx    \undefined \def \showISBNx     #1{\unskip}     \fi
\ifx \showISBNxiii \undefined \def \showISBNxiii  #1{\unskip}     \fi
\ifx \showISSN     \undefined \def \showISSN      #1{\unskip}     \fi
\ifx \showLCCN     \undefined \def \showLCCN      #1{\unskip}     \fi
\ifx \shownote     \undefined \def \shownote      #1{#1}          \fi
\ifx \showarticletitle \undefined \def \showarticletitle #1{#1}   \fi
\ifx \showURL      \undefined \def \showURL       {\relax}        \fi
% invisible to TeX
\providecommand\bibfield[2]{#2}
\providecommand\bibinfo[2]{#2}
\providecommand\natexlab[1]{#1}
\providecommand\showeprint[2][]{arXiv:#2}

\bibitem[\protect\citeauthoryear{Branke, Nguyen, Pickardt, and Zhang}{Branke
  et~al\mbox{.}}{2016}]%
        {Branke2016}
\bibfield{author}{\bibinfo{person}{Jürgen Branke}, \bibinfo{person}{Su
  Nguyen}, \bibinfo{person}{Christoph~W. Pickardt}, {and}
  \bibinfo{person}{Mengjie Zhang}.} \bibinfo{year}{2016}\natexlab{}.
\newblock \showarticletitle{Automated Design of Production Scheduling
  Heuristics: A Review}.
\newblock \bibinfo{journal}{\emph{IEEE Transactions on Evolutionary
  Computation}} \bibinfo{volume}{20}, \bibinfo{number}{1}
  (\bibinfo{year}{2016}), \bibinfo{pages}{110--124}.
\newblock
\urldef\tempurl%
\url{https://doi.org/10.1109/TEVC.2015.2429314}
\showDOI{\tempurl}


\bibitem[\protect\citeauthoryear{Dimopoulos and Zalzala}{Dimopoulos and
  Zalzala}{2001}]%
        {DIMOPOULOS2001}
\bibfield{author}{\bibinfo{person}{C. Dimopoulos} {and} \bibinfo{person}{A.M.S.
  Zalzala}.} \bibinfo{year}{2001}\natexlab{}.
\newblock \showarticletitle{Investigating the use of genetic programming for a
  classic one-machine scheduling problem}.
\newblock \bibinfo{journal}{\emph{Advances in Engineering Software}}
  \bibinfo{volume}{32}, \bibinfo{number}{6} (\bibinfo{year}{2001}),
  \bibinfo{pages}{489--498}.
\newblock
\showISSN{0965-9978}
\urldef\tempurl%
\url{https://doi.org/10.1016/S0965-9978(00)00109-5}
\showDOI{\tempurl}


\bibitem[\protect\citeauthoryear{Gil-Gala, Mencía, Sierra, and
  Varela}{Gil-Gala et~al\mbox{.}}{2020a}]%
        {Gil-Gala2020a}
\bibfield{author}{\bibinfo{person}{Francisco~J. Gil-Gala},
  \bibinfo{person}{Carlos Mencía}, \bibinfo{person}{María~R. Sierra}, {and}
  \bibinfo{person}{Ramiro Varela}.} \bibinfo{year}{2020}\natexlab{a}.
\newblock \showarticletitle{Learning ensembles of priority rules for online
  scheduling by hybrid evolutionary algorithms}.
\newblock \bibinfo{journal}{\emph{Integrated Computer-Aided Engineering}}
  \bibinfo{volume}{28}, \bibinfo{number}{1} (\bibinfo{date}{Dec.}
  \bibinfo{year}{2020}), \bibinfo{pages}{65–80}.
\newblock
\showISSN{10692509, 18758835}
\urldef\tempurl%
\url{https://doi.org/10.3233/ICA-200634}
\showDOI{\tempurl}


\bibitem[\protect\citeauthoryear{Gil-Gala, Sierra, Mencía, and
  Varela}{Gil-Gala et~al\mbox{.}}{2021}]%
        {GILGALA2021a}
\bibfield{author}{\bibinfo{person}{Francisco~J. Gil-Gala},
  \bibinfo{person}{María~R. Sierra}, \bibinfo{person}{Carlos Mencía}, {and}
  \bibinfo{person}{Ramiro Varela}.} \bibinfo{year}{2021}\natexlab{}.
\newblock \showarticletitle{Genetic programming with local search to evolve
  priority rules for scheduling jobs on a machine with time-varying capacity}.
\newblock \bibinfo{journal}{\emph{Swarm and Evolutionary Computation}}
  \bibinfo{volume}{66} (\bibinfo{year}{2021}), \bibinfo{pages}{100944}.
\newblock
\showISSN{2210-6502}
\urldef\tempurl%
\url{https://doi.org/10.1016/j.swevo.2021.100944}
\showDOI{\tempurl}


\bibitem[\protect\citeauthoryear{Gil-Gala, Sierra, Menc{\'{\i}}a, and
  Varela}{Gil-Gala et~al\mbox{.}}{2020b}]%
        {GilGala2020}
\bibfield{author}{\bibinfo{person}{Francisco~J. Gil-Gala},
  \bibinfo{person}{Mar{\'{\i}}a~R. Sierra}, \bibinfo{person}{Carlos
  Menc{\'{\i}}a}, {and} \bibinfo{person}{Ramiro Varela}.}
  \bibinfo{year}{2020}\natexlab{b}.
\newblock \showarticletitle{Combining hyper-heuristics to evolve ensembles of
  priority rules for on-line scheduling}.
\newblock \bibinfo{journal}{\emph{Natural Computing}} (\bibinfo{date}{June}
  \bibinfo{year}{2020}).
\newblock
\urldef\tempurl%
\url{https://doi.org/10.1007/s11047-020-09793-4}
\showDOI{\tempurl}


\bibitem[\protect\citeauthoryear{Gil-Gala and Varela}{Gil-Gala and
  Varela}{2019}]%
        {GilGala2019}
\bibfield{author}{\bibinfo{person}{Francisco~J. Gil-Gala} {and}
  \bibinfo{person}{Ramiro Varela}.} \bibinfo{year}{2019}\natexlab{}.
\newblock \showarticletitle{Genetic Algorithm to Evolve Ensembles of Rules for
  On-Line Scheduling on Single Machine with Variable Capacity}.
\newblock In \bibinfo{booktitle}{\emph{From Bioinspired Systems and Biomedical
  Applications to Machine Learning}}. \bibinfo{publisher}{Springer
  International Publishing}, \bibinfo{pages}{223--233}.
\newblock
\urldef\tempurl%
\url{https://doi.org/10.1007/978-3-030-19651-6_22}
\showDOI{\tempurl}


\bibitem[\protect\citeauthoryear{Hart and Sim}{Hart and Sim}{2016}]%
        {Hart2016}
\bibfield{author}{\bibinfo{person}{Emma Hart} {and} \bibinfo{person}{Kevin
  Sim}.} \bibinfo{year}{2016}\natexlab{}.
\newblock \showarticletitle{A Hyper-Heuristic Ensemble Method for Static
  Job-Shop Scheduling}.
\newblock \bibinfo{journal}{\emph{Evolutionary Computation}}
  \bibinfo{volume}{24}, \bibinfo{number}{4} (\bibinfo{year}{2016}),
  \bibinfo{pages}{609--635}.
\newblock
\urldef\tempurl%
\url{https://doi.org/10.1162/EVCO_a_00183}
\showDOI{\tempurl}


\bibitem[\protect\citeauthoryear{Jaklinović, Ðurasević, and
  Jakobović}{Jaklinović et~al\mbox{.}}{2021}]%
        {JAKLINOVIC2021}
\bibfield{author}{\bibinfo{person}{Kristijan Jaklinović},
  \bibinfo{person}{Marko Ðurasević}, {and} \bibinfo{person}{Domagoj
  Jakobović}.} \bibinfo{year}{2021}\natexlab{}.
\newblock \showarticletitle{Designing dispatching rules with genetic
  programming for the unrelated machines environment with constraints}.
\newblock \bibinfo{journal}{\emph{Expert Systems with Applications}}
  \bibinfo{volume}{172} (\bibinfo{year}{2021}), \bibinfo{pages}{114548}.
\newblock
\showISSN{0957-4174}
\urldef\tempurl%
\url{https://doi.org/10.1016/j.eswa.2020.114548}
\showDOI{\tempurl}


\bibitem[\protect\citeauthoryear{Jakobovic and Budin}{Jakobovic and
  Budin}{2006}]%
        {Jakobovic2006}
\bibfield{author}{\bibinfo{person}{Domagoj Jakobovic} {and}
  \bibinfo{person}{Leo Budin}.} \bibinfo{year}{2006}\natexlab{}.
\newblock \showarticletitle{Dynamic Scheduling with Genetic Programming}.
  \bibinfo{pages}{73--84}.
\newblock
\showISBNx{978-3-540-33143-8}
\urldef\tempurl%
\url{https://doi.org/10.1007/11729976_7}
\showDOI{\tempurl}


\bibitem[\protect\citeauthoryear{Masood, Chen, Mei, Al-Sahaf, and Zhang}{Masood
  et~al\mbox{.}}{2020}]%
        {Masood2020}
\bibfield{author}{\bibinfo{person}{Atiya Masood}, \bibinfo{person}{Gang Chen},
  \bibinfo{person}{Yi Mei}, \bibinfo{person}{Harith Al-Sahaf}, {and}
  \bibinfo{person}{Mengjie Zhang}.} \bibinfo{year}{2020}\natexlab{}.
\newblock \showarticletitle{A Fitness-based Selection Method for Pareto Local
  Search for Many-Objective Job Shop Scheduling}. In
  \bibinfo{booktitle}{\emph{2020 IEEE Congress on Evolutionary Computation
  (CEC)}}. \bibinfo{pages}{1--8}.
\newblock
\urldef\tempurl%
\url{https://doi.org/10.1109/CEC48606.2020.9185881}
\showDOI{\tempurl}


\bibitem[\protect\citeauthoryear{Miyashita}{Miyashita}{2000}]%
        {Miyashita2000}
\bibfield{author}{\bibinfo{person}{Kazuo Miyashita}.}
  \bibinfo{year}{2000}\natexlab{}.
\newblock \showarticletitle{Job-Shop Scheduling with Genetic Programming}. In
  \bibinfo{booktitle}{\emph{Proceedings of the 2nd Annual Conference on Genetic
  and Evolutionary Computation}} (Las Vegas, Nevada)
  \emph{(\bibinfo{series}{GECCO'00})}. \bibinfo{publisher}{Morgan Kaufmann
  Publishers Inc.}, \bibinfo{address}{San Francisco, CA, USA},
  \bibinfo{pages}{505–512}.
\newblock
\showISBNx{1558607080}


\bibitem[\protect\citeauthoryear{Nguyen, Mei, Xue, and Zhang}{Nguyen
  et~al\mbox{.}}{2019}]%
        {Nguyen2019}
\bibfield{author}{\bibinfo{person}{Su Nguyen}, \bibinfo{person}{Yi Mei},
  \bibinfo{person}{Bing Xue}, {and} \bibinfo{person}{Mengjie Zhang}.}
  \bibinfo{year}{2019}\natexlab{}.
\newblock \showarticletitle{{A Hybrid Genetic Programming Algorithm for
  Automated Design of Dispatching Rules}}.
\newblock \bibinfo{journal}{\emph{Evolutionary Computation}}
  \bibinfo{volume}{27}, \bibinfo{number}{3} (\bibinfo{date}{09}
  \bibinfo{year}{2019}), \bibinfo{pages}{467--496}.
\newblock
\showISSN{1063-6560}
\urldef\tempurl%
\url{https://doi.org/10.1162/evco_a_00230}
\showDOI{\tempurl}


\bibitem[\protect\citeauthoryear{Nguyen, Mei, and Zhang}{Nguyen
  et~al\mbox{.}}{2017a}]%
        {Nguyen2017}
\bibfield{author}{\bibinfo{person}{Su Nguyen}, \bibinfo{person}{Yi Mei}, {and}
  \bibinfo{person}{Mengjie Zhang}.} \bibinfo{year}{2017}\natexlab{a}.
\newblock \showarticletitle{Genetic programming for production scheduling: a
  survey with a unified framework}.
\newblock \bibinfo{journal}{\emph{Complex {\&} Intelligent Systems}}
  \bibinfo{volume}{3}, \bibinfo{number}{1} (\bibinfo{date}{Feb.}
  \bibinfo{year}{2017}), \bibinfo{pages}{41--66}.
\newblock
\urldef\tempurl%
\url{https://doi.org/10.1007/s40747-017-0036-x}
\showDOI{\tempurl}


\bibitem[\protect\citeauthoryear{Nguyen, Zhang, Johnston, and Tan}{Nguyen
  et~al\mbox{.}}{2013}]%
        {Nguyen2013}
\bibfield{author}{\bibinfo{person}{Su Nguyen}, \bibinfo{person}{Mengjie Zhang},
  \bibinfo{person}{Mark Johnston}, {and} \bibinfo{person}{Kay~Chen Tan}.}
  \bibinfo{year}{2013}\natexlab{}.
\newblock \showarticletitle{Dynamic Multi-objective Job Shop Scheduling: A
  Genetic Programming Approach}.
\newblock In \bibinfo{booktitle}{\emph{Studies in Computational Intelligence}}.
  \bibinfo{publisher}{Springer Berlin Heidelberg}, \bibinfo{pages}{251--282}.
\newblock
\urldef\tempurl%
\url{https://doi.org/10.1007/978-3-642-39304-4_10}
\showDOI{\tempurl}


\bibitem[\protect\citeauthoryear{Nguyen, Zhang, and Tan}{Nguyen
  et~al\mbox{.}}{2015}]%
        {Nguyen2015}
\bibfield{author}{\bibinfo{person}{Su Nguyen}, \bibinfo{person}{Mengjie Zhang},
  {and} \bibinfo{person}{Kay~Chen Tan}.} \bibinfo{year}{2015}\natexlab{}.
\newblock \showarticletitle{Enhancing genetic programming based
  hyper-heuristics for dynamic multi-objective job shop scheduling problems}.
  In \bibinfo{booktitle}{\emph{2015 IEEE Congress on Evolutionary Computation
  (CEC)}}. \bibinfo{pages}{2781--2788}.
\newblock
\urldef\tempurl%
\url{https://doi.org/10.1109/CEC.2015.7257234}
\showDOI{\tempurl}


\bibitem[\protect\citeauthoryear{Nguyen, Zhang, and Tan}{Nguyen
  et~al\mbox{.}}{2017b}]%
        {Nguyen2017a}
\bibfield{author}{\bibinfo{person}{Su Nguyen}, \bibinfo{person}{Mengjie Zhang},
  {and} \bibinfo{person}{Kay~Chen Tan}.} \bibinfo{year}{2017}\natexlab{b}.
\newblock \showarticletitle{Surrogate-Assisted Genetic Programming With
  Simplified Models for Automated Design of Dispatching Rules}.
\newblock \bibinfo{journal}{\emph{IEEE Transactions on Cybernetics}}
  \bibinfo{volume}{47}, \bibinfo{number}{9} (\bibinfo{year}{2017}),
  \bibinfo{pages}{2951--2965}.
\newblock
\urldef\tempurl%
\url{https://doi.org/10.1109/TCYB.2016.2562674}
\showDOI{\tempurl}


\bibitem[\protect\citeauthoryear{Park, Mei, Nguyen, Chen, Johnston, and
  Zhang}{Park et~al\mbox{.}}{2016}]%
        {Park2016}
\bibfield{author}{\bibinfo{person}{John Park}, \bibinfo{person}{Yi Mei},
  \bibinfo{person}{Su Nguyen}, \bibinfo{person}{Gang Chen},
  \bibinfo{person}{Mark Johnston}, {and} \bibinfo{person}{Mengjie Zhang}.}
  \bibinfo{year}{2016}\natexlab{}.
\newblock \showarticletitle{Genetic Programming Based Hyper-heuristics for
  Dynamic Job Shop Scheduling: Cooperative Coevolutionary Approaches}.
\newblock In \bibinfo{booktitle}{\emph{Lecture Notes in Computer Science}}.
  \bibinfo{publisher}{Springer International Publishing},
  \bibinfo{pages}{115--132}.
\newblock
\urldef\tempurl%
\url{https://doi.org/10.1007/978-3-319-30668-1_8}
\showDOI{\tempurl}


\bibitem[\protect\citeauthoryear{Park, Mei, Nguyen, Chen, and Zhang}{Park
  et~al\mbox{.}}{2017}]%
        {Park2017}
\bibfield{author}{\bibinfo{person}{John Park}, \bibinfo{person}{Yi Mei},
  \bibinfo{person}{Su Nguyen}, \bibinfo{person}{Gang Chen}, {and}
  \bibinfo{person}{Mengjie Zhang}.} \bibinfo{year}{2017}\natexlab{}.
\newblock \showarticletitle{An Investigation of Ensemble Combination Schemes
  for Genetic Programming based Hyper-heuristic Approaches to Dynamic Job Shop
  Scheduling}.
\newblock \bibinfo{journal}{\emph{Applied Soft Computing}}
  \bibinfo{volume}{63} (\bibinfo{date}{11} \bibinfo{year}{2017}).
\newblock
\urldef\tempurl%
\url{https://doi.org/10.1016/j.asoc.2017.11.020}
\showDOI{\tempurl}


\bibitem[\protect\citeauthoryear{Park, Mei, Nguyen, Chen, and Zhang}{Park
  et~al\mbox{.}}{2018}]%
        {Park2018a}
\bibfield{author}{\bibinfo{person}{John Park}, \bibinfo{person}{Yi Mei},
  \bibinfo{person}{Su Nguyen}, \bibinfo{person}{Gang Chen}, {and}
  \bibinfo{person}{Mengjie Zhang}.} \bibinfo{year}{2018}\natexlab{}.
\newblock \showarticletitle{Investigating a Machine Breakdown Genetic
  Programming Approach for Dynamic Job Shop Scheduling}.
\newblock In \bibinfo{booktitle}{\emph{Lecture Notes in Computer Science}}.
  \bibinfo{publisher}{Springer International Publishing},
  \bibinfo{pages}{253--270}.
\newblock
\urldef\tempurl%
\url{https://doi.org/10.1007/978-3-319-77553-1_16}
\showDOI{\tempurl}


\bibitem[\protect\citeauthoryear{Park, Nguyen, Zhang, and Johnston}{Park
  et~al\mbox{.}}{2015}]%
        {Park2015}
\bibfield{author}{\bibinfo{person}{John Park}, \bibinfo{person}{Su Nguyen},
  \bibinfo{person}{Mengjie Zhang}, {and} \bibinfo{person}{Mark Johnston}.}
  \bibinfo{year}{2015}\natexlab{}.
\newblock \showarticletitle{Evolving Ensembles of Dispatching Rules Using
  Genetic Programming for Job Shop Scheduling}. \bibinfo{pages}{92--104}.
\newblock
\showISBNx{978-3-319-16500-4}
\urldef\tempurl%
\url{https://doi.org/10.1007/978-3-319-16501-1_8}
\showDOI{\tempurl}


\bibitem[\protect\citeauthoryear{Pinedo}{Pinedo}{2012}]%
        {Pinedo2012}
\bibfield{author}{\bibinfo{person}{Michael~L. Pinedo}.}
  \bibinfo{year}{2012}\natexlab{}.
\newblock \bibinfo{booktitle}{\emph{Scheduling}}.
\newblock \bibinfo{publisher}{Springer {US}}.
\newblock
\urldef\tempurl%
\url{https://doi.org/10.1007/978-1-4614-2361-4}
\showDOI{\tempurl}


\bibitem[\protect\citeauthoryear{Planinić, {\DJ}urasević, and
  Jakobović}{Planinić et~al\mbox{.}}{2021}]%
        {Planinic2021}
\bibfield{author}{\bibinfo{person}{Lucija Planinić}, \bibinfo{person}{Marko
  {\DJ}urasević}, {and} \bibinfo{person}{Domagoj Jakobović}.}
  \bibinfo{year}{2021}\natexlab{}.
\newblock \showarticletitle{On the Application of $\epsilon$-Lexicase Selection
  in the Generation of Dispatching Rules}. In \bibinfo{booktitle}{\emph{2021
  IEEE Congress on Evolutionary Computation (CEC)}}.
  \bibinfo{pages}{2125--2132}.
\newblock
\urldef\tempurl%
\url{https://doi.org/10.1109/CEC45853.2021.9504982}
\showDOI{\tempurl}


\bibitem[\protect\citeauthoryear{Poli, Langdon, and McPhee}{Poli
  et~al\mbox{.}}{2008}]%
        {Poli2008}
\bibfield{author}{\bibinfo{person}{Riccardo Poli}, \bibinfo{person}{William~B.
  Langdon}, {and} \bibinfo{person}{Nicholas~Freitag McPhee}.}
  \bibinfo{year}{2008}\natexlab{}.
\newblock \bibinfo{booktitle}{\emph{A Field Guide to Genetic Programming}}.
\newblock \bibinfo{publisher}{Lulu Enterprises, UK Ltd}.
\newblock
\showISBNx{1409200736}


\bibitem[\protect\citeauthoryear{umi{\'{c}} and Jakobovi{\'{c}}}{umi{\'{c}} and
  Jakobovi{\'{c}}}{2021}]%
        {DUMIC2021107606}
\bibfield{author}{\bibinfo{person}{Mateja~\DJ umi{\'{c}}} {and}
  \bibinfo{person}{Domagoj Jakobovi{\'{c}}}.} \bibinfo{year}{2021}\natexlab{}.
\newblock \showarticletitle{Ensembles of priority rules for resource
  constrained project scheduling problem}.
\newblock \bibinfo{journal}{\emph{Applied Soft Computing}}
  \bibinfo{volume}{110} (\bibinfo{year}{2021}), \bibinfo{pages}{107606}.
\newblock
\showISSN{1568-4946}
\urldef\tempurl%
\url{https://doi.org/10.1016/j.asoc.2021.107606}
\showDOI{\tempurl}


\bibitem[\protect\citeauthoryear{Ðurasević and Jakobović}{Ðurasević and
  Jakobović}{2018}]%
        {DURASEVIC2018c}
\bibfield{author}{\bibinfo{person}{Marko Ðurasević} {and}
  \bibinfo{person}{Domagoj Jakobović}.} \bibinfo{year}{2018}\natexlab{}.
\newblock \showarticletitle{A survey of dispatching rules for the dynamic
  unrelated machines environment}.
\newblock \bibinfo{journal}{\emph{Expert Systems with Applications}}
  \bibinfo{volume}{113} (\bibinfo{year}{2018}), \bibinfo{pages}{555--569}.
\newblock
\showISSN{0957-4174}
\urldef\tempurl%
\url{https://doi.org/10.1016/j.eswa.2018.06.053}
\showDOI{\tempurl}


\bibitem[\protect\citeauthoryear{Đurasević and Jakobović}{Đurasević and
  Jakobović}{2020}]%
        {DURASEVIC2020}
\bibfield{author}{\bibinfo{person}{Marko Đurasević} {and}
  \bibinfo{person}{Domagoj Jakobović}.} \bibinfo{year}{2020}\natexlab{}.
\newblock \showarticletitle{Comparison of schedule generation schemes for
  designing dispatching rules with genetic programming in the unrelated
  machines environment}.
\newblock \bibinfo{journal}{\emph{Applied Soft Computing}}
  \bibinfo{volume}{96} (\bibinfo{year}{2020}), \bibinfo{pages}{106637}.
\newblock
\showISSN{1568-4946}
\urldef\tempurl%
\url{https://doi.org/10.1016/j.asoc.2020.106637}
\showDOI{\tempurl}


\bibitem[\protect\citeauthoryear{urasević, Jakobović, and
  Knežević}{urasević et~al\mbox{.}}{2016}]%
        {DURASEVIC2016}
\bibfield{author}{\bibinfo{person}{Marko~\DJ urasević},
  \bibinfo{person}{Domagoj Jakobović}, {and} \bibinfo{person}{Karlo
  Knežević}.} \bibinfo{year}{2016}\natexlab{}.
\newblock \showarticletitle{Adaptive scheduling on unrelated machines with
  genetic programming}.
\newblock \bibinfo{journal}{\emph{Applied Soft Computing}}
  \bibinfo{volume}{48} (\bibinfo{year}{2016}), \bibinfo{pages}{419--430}.
\newblock
\showISSN{1568-4946}
\urldef\tempurl%
\url{https://doi.org/10.1016/j.asoc.2016.07.025}
\showDOI{\tempurl}


\bibitem[\protect\citeauthoryear{{\DJ}urasevi{\'{c}} and
  Jakobovi{\'{c}}}{{\DJ}urasevi{\'{c}} and Jakobovi{\'{c}}}{2020}]%
        {Durasevi2020a}
\bibfield{author}{\bibinfo{person}{Marko {\DJ}urasevi{\'{c}}} {and}
  \bibinfo{person}{Domagoj Jakobovi{\'{c}}}.} \bibinfo{year}{2020}\natexlab{}.
\newblock \showarticletitle{Automatic design of dispatching rules for static
  scheduling conditions}.
\newblock \bibinfo{journal}{\emph{Neural Computing and Applications}}
  \bibinfo{volume}{33}, \bibinfo{number}{10} (\bibinfo{date}{Aug.}
  \bibinfo{year}{2020}), \bibinfo{pages}{5043--5068}.
\newblock
\urldef\tempurl%
\url{https://doi.org/10.1007/s00521-020-05292-w}
\showDOI{\tempurl}


\bibitem[\protect\citeauthoryear{urasevi{\'{c}} and
  Jakobovi{\'{c}}}{urasevi{\'{c}} and Jakobovi{\'{c}}}{2017a}]%
        {Durasevic2017a}
\bibfield{author}{\bibinfo{person}{Marko~\DJ urasevi{\'{c}}} {and}
  \bibinfo{person}{Domagoj Jakobovi{\'{c}}}.} \bibinfo{year}{2017}\natexlab{a}.
\newblock \showarticletitle{Comparison of ensemble learning methods for
  creating ensembles of dispatching rules for the unrelated machines
  environment}.
\newblock \bibinfo{journal}{\emph{Genetic Programming and Evolvable Machines}}
  \bibinfo{volume}{19}, \bibinfo{number}{1-2} (\bibinfo{date}{April}
  \bibinfo{year}{2017}), \bibinfo{pages}{53--92}.
\newblock
\urldef\tempurl%
\url{https://doi.org/10.1007/s10710-017-9302-3}
\showDOI{\tempurl}


\bibitem[\protect\citeauthoryear{urasevi{\'{c}} and
  Jakobovi{\'{c}}}{urasevi{\'{c}} and Jakobovi{\'{c}}}{2017b}]%
        {Durasevic2017}
\bibfield{author}{\bibinfo{person}{Marko~\DJ urasevi{\'{c}}} {and}
  \bibinfo{person}{Domagoj Jakobovi{\'{c}}}.} \bibinfo{year}{2017}\natexlab{b}.
\newblock \showarticletitle{Evolving dispatching rules for optimising
  many-objective criteria in the unrelated machines environment}.
\newblock \bibinfo{journal}{\emph{Genetic Programming and Evolvable Machines}}
  \bibinfo{volume}{19}, \bibinfo{number}{1-2} (\bibinfo{date}{Sept.}
  \bibinfo{year}{2017}), \bibinfo{pages}{9--51}.
\newblock
\urldef\tempurl%
\url{https://doi.org/10.1007/s10710-017-9310-3}
\showDOI{\tempurl}


\bibitem[\protect\citeauthoryear{urasevi{\'{c}} and
  Jakobovi{\'{c}}}{urasevi{\'{c}} and Jakobovi{\'{c}}}{2019}]%
        {Durasevic2019}
\bibfield{author}{\bibinfo{person}{Marko~\DJ urasevi{\'{c}}} {and}
  \bibinfo{person}{Domagoj Jakobovi{\'{c}}}.} \bibinfo{year}{2019}\natexlab{}.
\newblock \showarticletitle{Creating dispatching rules by simple ensemble
  combination}.
\newblock \bibinfo{journal}{\emph{Journal of Heuristics}} \bibinfo{volume}{25},
  \bibinfo{number}{6} (\bibinfo{date}{May} \bibinfo{year}{2019}),
  \bibinfo{pages}{959--1013}.
\newblock
\urldef\tempurl%
\url{https://doi.org/10.1007/s10732-019-09416-x}
\showDOI{\tempurl}


\bibitem[\protect\citeauthoryear{Vlašić, Ðurasević, and
  Jakobović}{Vlašić et~al\mbox{.}}{2019}]%
        {VLASIC2019}
\bibfield{author}{\bibinfo{person}{Ivan Vlašić}, \bibinfo{person}{Marko
  Ðurasević}, {and} \bibinfo{person}{Domagoj Jakobović}.}
  \bibinfo{year}{2019}\natexlab{}.
\newblock \showarticletitle{Improving genetic algorithm performance by
  population initialisation with dispatching rules}.
\newblock \bibinfo{journal}{\emph{Computers \& Industrial Engineering}}
  \bibinfo{volume}{137} (\bibinfo{year}{2019}), \bibinfo{pages}{106030}.
\newblock
\showISSN{0360-8352}
\urldef\tempurl%
\url{https://doi.org/10.1016/j.cie.2019.106030}
\showDOI{\tempurl}


\bibitem[\protect\citeauthoryear{Wang, Mei, Park, and Zhang}{Wang
  et~al\mbox{.}}{2019a}]%
        {Wang2019}
\bibfield{author}{\bibinfo{person}{Shaolin Wang}, \bibinfo{person}{Yi Mei},
  \bibinfo{person}{John Park}, {and} \bibinfo{person}{Mengjie Zhang}.}
  \bibinfo{year}{2019}\natexlab{a}.
\newblock \showarticletitle{Evolving Ensembles of Routing Policies using
  Genetic Programming for Uncertain Capacitated Arc Routing Problem}. In
  \bibinfo{booktitle}{\emph{2019 IEEE Symposium Series on Computational
  Intelligence (SSCI)}}. \bibinfo{pages}{1628--1635}.
\newblock
\urldef\tempurl%
\url{https://doi.org/10.1109/SSCI44817.2019.9002749}
\showDOI{\tempurl}


\bibitem[\protect\citeauthoryear{Wang, Mei, Park, and Zhang}{Wang
  et~al\mbox{.}}{2019b}]%
        {Wang2019a}
\bibfield{author}{\bibinfo{person}{Shaolin Wang}, \bibinfo{person}{Yi Mei},
  \bibinfo{person}{John Park}, {and} \bibinfo{person}{Mengjie Zhang}.}
  \bibinfo{year}{2019}\natexlab{b}.
\newblock \showarticletitle{Evolving Ensembles of Routing Policies using
  Genetic Programming for Uncertain Capacitated Arc Routing Problem}. In
  \bibinfo{booktitle}{\emph{2019 IEEE Symposium Series on Computational
  Intelligence (SSCI)}}. \bibinfo{pages}{1628--1635}.
\newblock
\urldef\tempurl%
\url{https://doi.org/10.1109/SSCI44817.2019.9002749}
\showDOI{\tempurl}


\bibitem[\protect\citeauthoryear{Wolpert and Macready}{Wolpert and
  Macready}{1997}]%
        {Wolpert1997}
\bibfield{author}{\bibinfo{person}{D.H. Wolpert} {and} \bibinfo{person}{W.G.
  Macready}.} \bibinfo{year}{1997}\natexlab{}.
\newblock \showarticletitle{No free lunch theorems for optimization}.
\newblock \bibinfo{journal}{\emph{IEEE Transactions on Evolutionary
  Computation}} \bibinfo{volume}{1}, \bibinfo{number}{1}
  (\bibinfo{year}{1997}), \bibinfo{pages}{67--82}.
\newblock
\urldef\tempurl%
\url{https://doi.org/10.1109/4235.585893}
\showDOI{\tempurl}


\bibitem[\protect\citeauthoryear{Zhang, Mei, Nguyen, Tan, and Zhang}{Zhang
  et~al\mbox{.}}{2021c}]%
        {Zhang2021}
\bibfield{author}{\bibinfo{person}{Fangfang Zhang}, \bibinfo{person}{Yi Mei},
  \bibinfo{person}{Su Nguyen}, \bibinfo{person}{Kay~Chen Tan}, {and}
  \bibinfo{person}{Mengjie Zhang}.} \bibinfo{year}{2021}\natexlab{c}.
\newblock \showarticletitle{Multitask Genetic Programming-Based Generative
  Hyperheuristics: A Case Study in Dynamic Scheduling}.
\newblock \bibinfo{journal}{\emph{IEEE Transactions on Cybernetics}}
  (\bibinfo{year}{2021}), \bibinfo{pages}{1--14}.
\newblock
\urldef\tempurl%
\url{https://doi.org/10.1109/TCYB.2021.3065340}
\showDOI{\tempurl}


\bibitem[\protect\citeauthoryear{Zhang, Mei, Nguyen, and Zhang}{Zhang
  et~al\mbox{.}}{2020}]%
        {Zhang2020c}
\bibfield{author}{\bibinfo{person}{Fangfang Zhang}, \bibinfo{person}{Yi Mei},
  \bibinfo{person}{Su Nguyen}, {and} \bibinfo{person}{Mengjie Zhang}.}
  \bibinfo{year}{2020}\natexlab{}.
\newblock \showarticletitle{Guided Subtree Selection for Genetic Operators in
  Genetic Programming for Dynamic Flexible Job Shop Scheduling}.
\newblock In \bibinfo{booktitle}{\emph{Lecture Notes in Computer Science}}.
  \bibinfo{publisher}{Springer International Publishing},
  \bibinfo{pages}{262--278}.
\newblock
\urldef\tempurl%
\url{https://doi.org/10.1007/978-3-030-44094-7_17}
\showDOI{\tempurl}


\bibitem[\protect\citeauthoryear{Zhang, Mei, Nguyen, and Zhang}{Zhang
  et~al\mbox{.}}{2021a}]%
        {Zhang2021c}
\bibfield{author}{\bibinfo{person}{Fangfang Zhang}, \bibinfo{person}{Yi Mei},
  \bibinfo{person}{Su Nguyen}, {and} \bibinfo{person}{Mengjie Zhang}.}
  \bibinfo{year}{2021}\natexlab{a}.
\newblock \showarticletitle{Collaborative Multifidelity-Based Surrogate Models
  for Genetic Programming in Dynamic Flexible Job Shop Scheduling}.
\newblock \bibinfo{journal}{\emph{IEEE Transactions on Cybernetics}}
  (\bibinfo{year}{2021}), \bibinfo{pages}{1--15}.
\newblock
\urldef\tempurl%
\url{https://doi.org/10.1109/TCYB.2021.3050141}
\showDOI{\tempurl}


\bibitem[\protect\citeauthoryear{Zhang, Mei, Nguyen, and Zhang}{Zhang
  et~al\mbox{.}}{2021b}]%
        {Zhang2021d}
\bibfield{author}{\bibinfo{person}{Fangfang Zhang}, \bibinfo{person}{Yi Mei},
  \bibinfo{person}{Su Nguyen}, {and} \bibinfo{person}{Mengjie Zhang}.}
  \bibinfo{year}{2021}\natexlab{b}.
\newblock \showarticletitle{Evolving Scheduling Heuristics via Genetic
  Programming With Feature Selection in Dynamic Flexible Job-Shop Scheduling}.
\newblock \bibinfo{journal}{\emph{IEEE Transactions on Cybernetics}}
  \bibinfo{volume}{51}, \bibinfo{number}{4} (\bibinfo{year}{2021}),
  \bibinfo{pages}{1797--1811}.
\newblock
\urldef\tempurl%
\url{https://doi.org/10.1109/TCYB.2020.3024849}
\showDOI{\tempurl}


\bibitem[\protect\citeauthoryear{Zhang, Mei, Nguyen, Zhang, and Tan}{Zhang
  et~al\mbox{.}}{2021d}]%
        {Zhang2021a}
\bibfield{author}{\bibinfo{person}{Fangfang Zhang}, \bibinfo{person}{Yi Mei},
  \bibinfo{person}{Su Nguyen}, \bibinfo{person}{Mengjie Zhang}, {and}
  \bibinfo{person}{Kay~Chen Tan}.} \bibinfo{year}{2021}\natexlab{d}.
\newblock \showarticletitle{Surrogate-Assisted Evolutionary Multitask Genetic
  Programming for Dynamic Flexible Job Shop Scheduling}.
\newblock \bibinfo{journal}{\emph{IEEE Transactions on Evolutionary
  Computation}} \bibinfo{volume}{25}, \bibinfo{number}{4}
  (\bibinfo{year}{2021}), \bibinfo{pages}{651--665}.
\newblock
\urldef\tempurl%
\url{https://doi.org/10.1109/TEVC.2021.3065707}
\showDOI{\tempurl}


\bibitem[\protect\citeauthoryear{Zhang, Mei, and Zhang}{Zhang
  et~al\mbox{.}}{2019a}]%
        {Zhang2019}
\bibfield{author}{\bibinfo{person}{Fangfang Zhang}, \bibinfo{person}{Yi Mei},
  {and} \bibinfo{person}{Mengjie Zhang}.} \bibinfo{year}{2019}\natexlab{a}.
\newblock \showarticletitle{Evolving Dispatching Rules for Multi-objective
  Dynamic Flexible Job Shop Scheduling via Genetic Programming
  Hyper-heuristics}. In \bibinfo{booktitle}{\emph{2019 IEEE Congress on
  Evolutionary Computation (CEC)}}. \bibinfo{pages}{1366--1373}.
\newblock
\urldef\tempurl%
\url{https://doi.org/10.1109/CEC.2019.8790112}
\showDOI{\tempurl}


\bibitem[\protect\citeauthoryear{Zhang, Mei, and Zhang}{Zhang
  et~al\mbox{.}}{2019b}]%
        {Zhang2019d}
\bibfield{author}{\bibinfo{person}{Fangfang Zhang}, \bibinfo{person}{Yi Mei},
  {and} \bibinfo{person}{Mengjie Zhang}.} \bibinfo{year}{2019}\natexlab{b}.
\newblock \showarticletitle{A Two-Stage Genetic Programming Hyper-Heuristic
  Approach with Feature Selection for Dynamic Flexible Job Shop Scheduling}. In
  \bibinfo{booktitle}{\emph{Proceedings of the Genetic and Evolutionary
  Computation Conference}} (Prague, Czech Republic)
  \emph{(\bibinfo{series}{GECCO '19})}. \bibinfo{publisher}{Association for
  Computing Machinery}, \bibinfo{address}{New York, NY, USA},
  \bibinfo{pages}{347–355}.
\newblock
\showISBNx{9781450361118}
\urldef\tempurl%
\url{https://doi.org/10.1145/3321707.3321790}
\showDOI{\tempurl}


\end{thebibliography}

\end{document}